\documentclass{article}

\usepackage{arxiv}

\usepackage[utf8]{inputenc} 
\usepackage[T1]{fontenc}    
\usepackage{hyperref}       
\usepackage{url}            
\usepackage{booktabs}       
\usepackage{nicefrac}       
\usepackage{microtype}      
\usepackage{xcolor}
\usepackage{lipsum}
\usepackage{graphicx}
\usepackage{amsmath,amssymb,amsfonts}
\usepackage{algorithmic}
\usepackage{textcomp}
\usepackage{booktabs}
\usepackage{scalerel}
\usepackage{array}
\newcolumntype{C}[1]{>{\centering\arraybackslash}m{#1}}
\graphicspath{ {./images/} }

\title{Unifying Scale-Aware Depth Prediction and Perceptual Priors for Monocular Endoscope Pose Estimation and Tissue Reconstruction}

\author{
  \parbox{\linewidth}{\centering
    Muzammil Khan$^{1}$\thanks{Preprint Notice - This work has been submitted to \textit{IEEE Access} for possible publication. This is the author’s original manuscript and has not undergone peer review. Subsequent versions of this manuscript may differ from this one. The final published version, if accepted, will be available via IEEE Xplore. Corresponding author: \texttt{m.khan@utwente.nl}} \hfill
    Enzo Kerkhof$^{2}$ \hfill
    Matteo Fusaglia$^{2}$ \\[0.8em] 
    Koert Kuhlmann$^{2}$ \hfill
    Theo Ruers$^{2}$ \hfill
    Fran\c{c}oise J. Siepel$^{1}$
  }
  \AND
  \normalfont
  $^{1}$The Robotics and Mechatronics Group, University of Twente, \\ Drienerlolaan 5, 7522 NB Enschede, The Netherlands \\[0.5em]
  $^{2}$The Department of Surgical Oncology, The Netherlands Cancer Institute-Antoni van Leeuwenhoek, \\ Plesmanlaan 121, 1066 CX, Amsterdam, The Netherlands \\[0.5em]
  \texttt{m.khan@utwente.nl}, \texttt{e.kerkhof@nki.nl}, \texttt{m.fusaglia@nki.nl}, \\ \texttt{k.kuhlmann@nki.nl}, \texttt{t.ruers@nki.nl}, \texttt{f.j.siepel@utwente.nl}
}

\begin{document}
\maketitle
\begin{abstract}
 Accurate endoscope pose estimation and \textit{3D} tissue surface reconstruction significantly enhances monocular minimally invasive surgical procedures by enabling accurate navigation and improved spatial awareness. However, monocular endoscope pose estimation and tissue reconstruction face persistent challenges, including depth ambiguity, physiological tissue deformation, inconsistent endoscope motion, limited texture fidelity, and a restricted field of view. To overcome these limitations, a unified framework for monocular endoscopic tissue reconstruction that integrates scale-aware depth prediction with temporally-constrained perceptual refinement is presented. This framework incorporates a novel MAPIS-Depth module, which leverages Depth Pro for robust initialisation and Depth Anything for efficient per-frame depth prediction, in conjunction with L-BFGS-B optimisation, to generate pseudo-metric depth estimates. These estimates are temporally refined by computing pixel correspondences using RAFT and adaptively blending flow-warped frames based on LPIPS perceptual similarity, thereby reducing artefacts arising from physiological tissue deformation and motion. To ensure accurate registration of the synthesised pseudo-RGBD frames from MAPIS-Depth, a novel WEMA-RTDL module is integrated, optimising both rotation and translation. Finally, truncated signed distance function-based volumetric fusion and marching cubes are applied to extract a comprehensive \textit{3D} surface mesh. Evaluations on HEVD and SCARED, with ablation and comparative analyses, demonstrate the framework’s robustness and superiority over state-of-the-art methods.
\end{abstract}

\keywords{Monocular reconstruction \and Depth estimation \and Endoscope pose estimation \and Foundation model \and Monocular endoscopy.}

  \section{Introduction}
 \label{sec:intro}
 Minimally invasive surgery (MIS), including monocular and stereoscopic endoscopy as well as robotic systems, has transformed modern surgical practice by reducing incision size, limiting tissue trauma and lowering postoperative morbidity and recovery times~\cite{birkett1995three, collins2012towards}. Among these approaches, monocular minimally invasive surgery (MMIS) remains the most widely adopted technique due to its lower cost, simpler setup, and broad accessibility across diverse clinical settings~\cite{hodgson1999effects, limb2023principles}. However, MMIS is inherently constrained by a limited field of view and the absence of direct depth cues, forcing surgeons to rely on indirect visual indicators such as shading, motion, or anatomical landmarks to interpret spatial relationships. These perceptual limitations are further exacerbated in anatomically complex regions such as those involving hidden tumours, overlapping tissues, or proximity to critical vessels, posing challenges to accurate navigation and intraoperative decision-making. Whilst stereoscopic and robotic systems partially alleviate these issues through binocular disparity or enhanced visual interfaces, their adoption remains limited due to higher costs and integration complexities. In this context, recent advances in computer vision and deep learning have enabled computational techniques that can infer depth and spatial structure directly from monocular endoscopic images~\cite{liu2023surface, wang2024non, hayoz2023learning, wu20223d}. These approaches offer a hardware-agnostic solution to enhance spatial awareness and intraoperative guidance within existing MMIS workflows, representing a significant step towards safer, more precise, and context-aware MIS procedures.
 \par
 Despite these advancements, achieving accurate \textit{3D} surface reconstruction in MMIS remains technically challenging, primarily due to the lack of metric depth cues, which are essential for precise spatial measurements and accuracy in navigation~\cite{verri1986absolute, shao2016models}. Monocular systems lack stereoscopic disparity, relying instead on secondary visual cues such as motion parallax, shading gradients, and occlusion boundaries~\cite{arampatzakis2023monocular, su2023opa, yoo2023monocular, hibbard2023luminance}. However, these inferred depth cues often result in non-metric or relative depth estimates, which are prone to noise and heavily dependent on the structural properties of the scene. Furthermore, the MMIS environment introduces dynamic, non-rigid deformations due to physiological processes such as respiration, complicating robust depth estimation and temporal alignment. These challenges significantly hinder the ability to achieve consistent metric depth accuracy~\cite{liu2023surface, wang2024non, hayoz2023learning}.
 \par
 Another major challenge in \textit{3D} reconstruction is the reliable registration of RGB-D frames, particularly in low-texture, homogeneous tissue regions commonly encountered in MMIS. The scarcity of distinctive features in these regions impairs the identification of stable spatial reference points, which are critical for effective feature matching and alignment~\cite{wu20223d, wang2023monocular}. This issue is further exacerbated in complex anatomical environments, where the endoscope must navigate through narrow, curved, and multi-layered pathways. These constraints necessitate high-precision depth and endoscope pose estimation strategies to ensure accurate and stable reconstructions~\cite{allan2021stereo, bobrow2023colonoscopy}.

  \subsection{Contribution}
 The primary contributions of this research are encapsulated in the design and development of a unified framework for monocular endoscope pose estimation and tissue reconstruction, which integrates scale-aware depth prediction with perceptually guided refinement and leverages both data-driven and optimisation-based methodologies. It comprises the following three core modules:
 \begin{itemize}
 
 \item \textbf{Metric-Aware Perceptual Image Similarity Depth (MAPIS-Depth) module:} A novel depth estimation module that combines the strengths of the foundation models Depth Pro~\cite{bochkovskii2024depth} and Depth Anything~\cite{yang2024depth}, the former used for robust initialisation and the latter for efficient per-frame inference. The resulting depth maps are further refined using L-BFGS-B optimisation. To enhance temporal consistency and suppress artefacts from tissue deformation and motion, RAFT-based optical flow~\cite{teed2020raft} is used to establish inter-frame correspondences, and LPIPS (Learned Perceptual Image Patch Similarity)~\cite{zhang2018unreasonable} is employed to guide adaptive perceptual blending.
 
 \item \textbf{Weighted Exponential Moving Average-based Rotation-Translation Dog-Leg (WEMA-RTDL) module:} A robust, multiscale $\mathcal{SO}(3)$–$\mathcal{SE}(3)$ decoupled dog-leg optimisation strategy~\cite{makaev2019software}, developed to improve frame-to-frame endoscope pose estimation in low-texture, homogeneous regions. It incorporates a WEMA refinement mechanism~\cite{makaev2019software, brown1956exponential} to ensure stable and accurate transformation estimates.
 
 \item \textbf{Volumetric fusion module:} A truncated signed distance function (TSDF)-based~\cite{tsdfOg} approach for smooth monocular tissue surface reconstruction, with Marching Cubes \cite{lorensen1998marching} used to extract a triangulated mesh \cite{recasens2021endo}.

 \end{itemize}

 Furthermore, the proposed framework is validated across multiple benchmark datasets, including Hamlyn Endoscopic Video Dataset (HEVD) \cite{recasens2021endo} and Stereo Correspondence and Reconstruction of Endoscopic Data (SCARED)~\cite{allan2021stereo}, with comprehensive performance assessments conducted through ablation and comparative studies.

   \section{Related Works \label{relatedwork}}

 Recent years have witnessed significant progress in monocular surface reconstruction, driven by advances in both learning-based and optimisation-based methodologies~\cite{samavati2023deep, farshian2023deep, murai2024mast3r, matsuki2024gaussian, fan2024instantsplat}. These approaches can be broadly categorised into offline and online methods, with online reconstruction being essential for real-time applications such as MMIS procedures. Online techniques typically involve real-time depth estimation, spatial correspondence modelling, and surface integration to produce temporally consistent monocular reconstructions~\cite{moravec1980obstacle, soares2021crowd}. These techniques are further categorised into geometry-driven, end-to-end deep learning-based, and hybrid approaches, each offering distinct advantages in balancing computational efficiency, accuracy, and adaptability to complex environments~\cite{hsiao2013study, blanz2004statistical, lhuillier2005quasi, dall2014comparison, park2005hybrid}.

 \subsection{Geometry-driven techniques}
 Classical methods for monocular reconstruction commonly rely on geometric techniques involving feature detection, matching, and camera motion estimation~\cite{taketomi2017visual, jeong2006visual}. Algorithms such as ORB-SLAM~\cite{mur2015orb}, ORB-SLAM3~\cite{campos2021orb}, and PTAM~\cite{castle2010parallel} employ local descriptors including SIFT, SURF, and ORB~\cite{lowe1999object, bay2008speeded, rublee2011orb}, combined with pose optimisation via Perspective-n-Point solvers and bundle adjustment~\cite{fischler1981random, triggs2000bundle}. Probabilistic formulations such as~\cite{grasa2011ekf, martinez2005unscented, holmes2008square} have improved robustness to motion uncertainty, however, these approaches often produce sparse reconstructions and underperform in texture-sparse or low-contrast environments typical in MMIS. Variational methods have improved map density~\cite{zhang20233d, li2020super, mahmoud2018live, wang2023advanced, zhao2023spsvo}, yet generally lack depth scale estimation and assume rigid scene structure, limitations that are particularly pronounced in anatomically dynamic surgical settings.

 \subsection{End-to-end deep learning-based techniques}

 Deep learning-based techniques have transformed monocular reconstruction by unifying depth prediction, feature extraction, and motion estimation within a single trainable framework~\cite{zhou2017unsupervised, li2018ongoing}. Unlike traditional methods that depend on hand-crafted features, these approaches leverage convolutional and recurrent neural networks to infer depth and pose directly from image sequences~\cite{li2017indoor, clark2017vidloc}. PoseNet~\cite{kendall2015posenet} and DeepVO~\cite{wang2017deepvo} exemplify architectures that regress camera motion, while models such as SfMLearner~\cite{li2018ongoing} and UnDeepVO~\cite{li2018undeepvo} jointly estimate depth and ego-motion, often with self-supervised loss formulations. Other works like NICESLAM~\cite{zhu2022nice} utilise hierarchical neural fields for dense reconstruction, and Monodepth~\cite{mondepth} demonstrates how dense disparity can be recovered from monocular inputs. However, the effectiveness of these methods in MMIS contexts is often constrained by their reliance on large-scale labelled datasets, which are difficult to obtain in medical domains~\cite{hashimoto2018artificial}. This underscores the need for data-efficient solutions that can generalise across anatomical variations.

 \subsection{Hybrid techniques}
 To overcome the limitations of purely geometric or learning-based methods, recent research has explored hybrid frameworks that combine learning-driven depth inference with geometric refinement and surface integration. For example, Endo-Depth-and-Motion~\cite{recasens2021endo} integrates Monodepth2-based depth prediction with photometric consistency for motion estimation, followed by volumetric fusion via TSDF~\cite{tsdfOg}. Similarly, EndoGSLAM~\cite{wang2024endogslam} applies \textit{3D} Gaussian splatting and differentiable rasterisation~\cite{kerbl20233d} to reconstruct detailed anatomical surfaces in real time. It includes both high-fidelity and fast variants, with pre-filtering mechanisms to exclude unreliable pixel regions and improve temporal coherence. Other modular designs such as BodySLAM~\cite{manni2024bodyslam} incorporate deep depth networks like ZoeDepth~\cite{bhat2023zoedepth} for monocular depth estimation and CyclePose for pose estimation, fusing these with traditional techniques such as pose graph optimisation and ICP~\cite{li2021globally}.
 \par
 Overall, these works highlight the growing trend towards integrating perceptual depth cues, data-driven learning, and geometric consistency for robust and scalable monocular reconstruction, especially under the constraints of tissue deformation, low texture, and scale ambiguity found in MMIS environments.

   \section{Methodology}
 \label{sec:formatting}
 The proposed framework integrates three core modules: MAPIS-Depth, which generates scale-consistent depth estimates; WEMA-RTDL, a robust pose estimation module for registering pseudo-RGBD frames; and a volumetric fusion module, which enables high-resolution tissue surface reconstruction.

 \subsection{MAPIS-Depth Module}
\begin{figure*}
     \centering
     \includegraphics[width=1.0\linewidth]{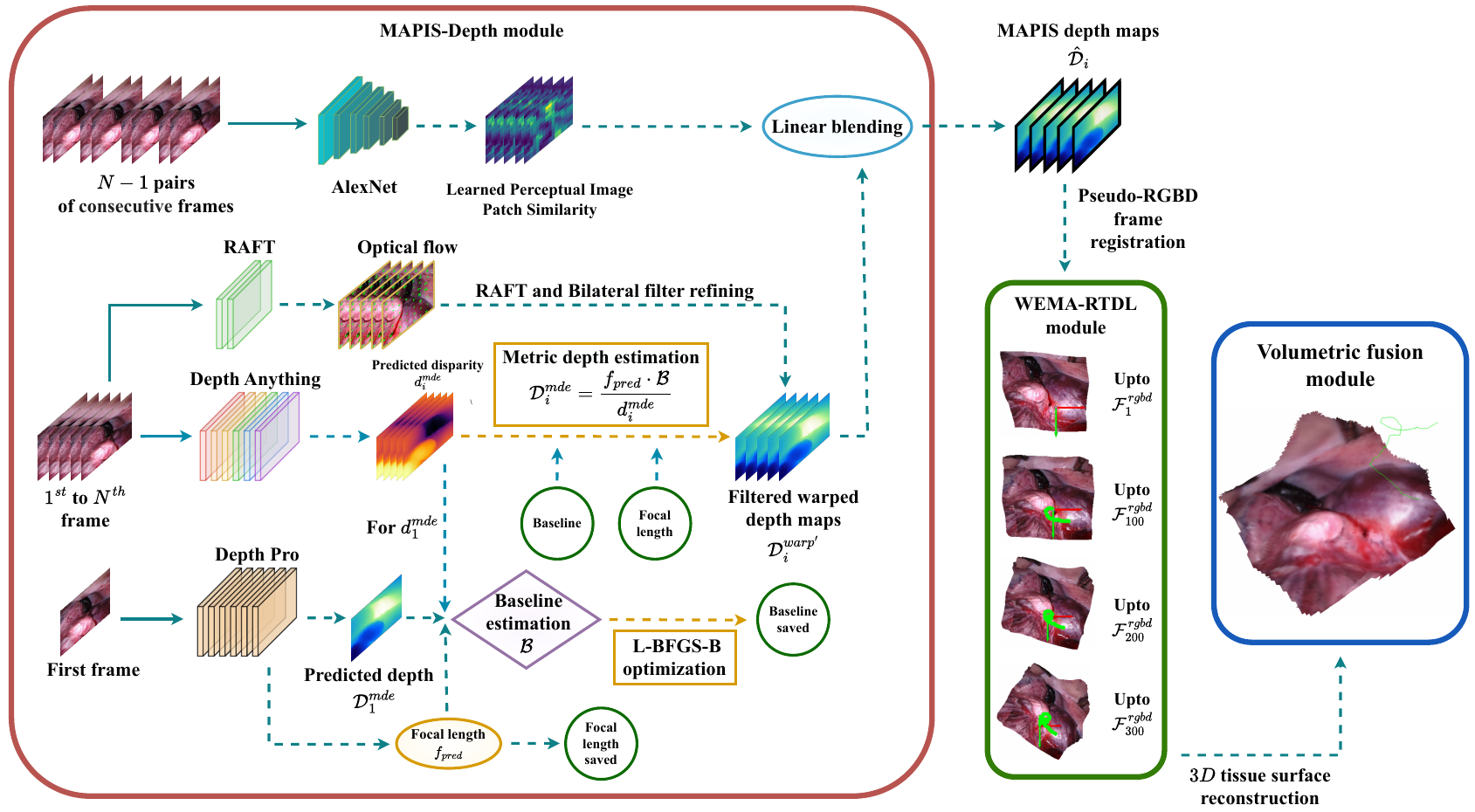}
     \caption{Methodology of the proposed framework with three modules: MAPIS-Depth, WEMA-RTDL, and Volumetric Fusion.}
     \label{depthest}
 \end{figure*}

 The MAPIS-Depth module, illustrated in Fig.~\ref{depthest}, plays a central role in the proposed framework by producing high-fidelity scale-consistent depth maps essential for accurate tissue reconstruction. It leverages two vision transformer-based foundation models, Depth Pro~\cite{bochkovskii2024depth} and Depth Anything~\cite{yang2024depth}, to capture complex spatial hierarchies and structural variations typical of MMIS environments~\cite{han2022survey}. Depth Pro facilitates zero-shot metric depth prediction with precise boundary delineation through multi-scale attention, while Depth Anything provides rapid per-frame inference and improved robustness to specular highlights and visual distortions, both of which are prevalent in endoscopic imaging.
 \par
 To optimise inference speed while maintaining scale consistency, Depth Pro is employed solely for the initialisation stage, wherein it estimates both the depth and the corresponding focal length for the first frame in the input sequence. For subsequent frames, depth is inferred using the more computationally efficient Depth Anything model. Specifically, the first frame, $\mathcal{J}_{1}$, is processed by Depth Pro to estimate its corresponding pseudo-metric depth estimate (MDE) $\mathcal{D}_{1}^{mde}$ and a focal length $f_{pred}$. Simultaneously, the same frame is input into Depth Anything, which predicts the associated disparity map $d_{1}^{mde}$. To enable scale-aware depth estimation for the remaining frames, a pseudo-stereo baseline $\mathcal{B}$ is estimated from the initial frame by utilising $\mathcal{D}_{1}^{mde}$, $d_{1}^{mde}$, and $f_{pred}$. This baseline represents the effective distance between two virtual camera centres and emulates a stereo configuration within a monocular context. The optimal value of $\mathcal{B}$ is computed by minimising the discrepancy between the known depth $\mathcal{D}_{1}^{mde}$ and the depth derived from the disparity map, as follows:
 \begin{equation}
     \mathcal{B} = \arg \min_{\hat{\mathcal{B}}} \left(\mathcal{D}_{1}^{mde} - \mathcal{D}_{1}^{est}  \right)^{2}
 \end{equation}

 where $\mathcal{D}_{1}^{est} = \frac{f_{pred} \cdot \hat{\mathcal{B}}}{d_{1}^{mde}}$ represents the computed depth, and $\hat{\mathcal{B}}$ denotes the initial estimate of the baseline. The objective function is minimised using the L-BFGS-B algorithm~\cite{zhu1997algorithm}, which iteratively refines $\hat{\mathcal{B}}$ to yield the optimal baseline $\mathcal{B}$. Once $\mathcal{B}$ is determined, the subsequent frames are directly processed by the Depth Anything model to predict their respective disparity maps $\{ d_{i}^{mde} \}_{i=2}^{N}$, where $N$ denotes the total number of frames. The corresponding MDEs $\{ \mathcal{D}_{i}^{mde} \}_{i=2}^{N}$ are then computed using the relation $\mathcal{D}_{i}^{mde} = \frac{f_{pred} \cdot \mathcal{B}}{d_{i}^{mde}}$.
 \par
 Up to this point, the process treats each consecutive image pair, $\mathcal{J}{i-1}$ and $\mathcal{J}{i}$, as independent snapshots. However, this approach overlooks the temporal dynamics inherent in MMIS sequences, including physiological tissue deformations and endoscope movement. To compensate for this, the sequence $\{\mathcal{D}_{i}^{mde}\}_{i=1}^{N}$, containing $N$ MDEs corresponding to the input image sequence $\{\mathcal{J}_{i}\}_{i=1}^{N}$, is refined using optical flow analysis and LPIPS maps. For each pair of consecutive frames $\{ \mathcal{J}_{i-1}, \mathcal{J}_{i} \}$, dense optical flow is computed to estimate pixel-wise displacements, which map positions from $\mathcal{J}_{i-1}$ to their correspondences in $\mathcal{J}_{i}$. This analysis is performed using the Recurrent All-Pairs Field Transforms (RAFT) method~\cite{teed2020raft}, which yields high-resolution flow fields with precise pixel-level accuracy. Its use of multi-scale correlation volumes enables robust flow estimation, which is critical for achieving consistent temporal alignment of depth predictions.
 \par
 Once the optical flow $(\mathcal{U}, \mathcal{V})$ is computed, it is used to warp $\mathcal{D}_{i-1}^{mde}$ using the \textit{remap} function in \textit{OpenCV}~\cite{opencv_library}, producing a warped depth map, $\mathcal{D}_{i}^{warp}$. Furthermore, to preserve the structural integrity of the depth map and ensure that key features are not lost during the warping process, a bilateral filtering~\cite{tomasi1998bilateral} step is applied to $\mathcal{D}_{i}^{warp}$ as follows:
 \begin{equation}
 \begin{split}
 \mathcal{D}_{I}^{warp'}(\mathbf{x}) = & \frac{1}{W(\mathbf{x})} \sum_{\mathbf{y} \in \Omega} \mathcal{D}_{i}^{warp}(\mathbf{y}) \cdot \exp \left( -\frac{\|\mathbf{x} - \mathbf{y}\|^2}{2\sigma_d^2} \right) \\ & \cdot \exp \left( -\frac{|\mathcal{D}_{i}^{warp}(\mathbf{x}) - \mathcal{D}_{i}^{warp}(\mathbf{y})|^2}{2\sigma_r^2} \right)
 \end{split}
 \end{equation}
 where $\mathcal{D}_{i}^{warp'}$ is the filtered warped depth map, $W(\mathbf{x})$ is the normalisation term, $\Omega$ is the neighbourhood of $\mathbf{x}$, with $\mathbf{x}$ and $\mathbf{y}$ being the pixel indices in $\Omega$. The parameters $\sigma_{d}$ and $\sigma_{r}$ control the spatial extent and the range of intensity considered for smoothing, respectively.
 \par
 Additionally, the image pair $\{ \mathcal{J}_{i-1}, \mathcal{J}_{i} \}$ from the sequence $\{ \mathcal{J}_{i} \}_{i=1}^{N}$ is input into a pre-trained AlexNet model~\cite{krizhevsky2017imagenet}, which has been trained on the ImageNet dataset with its classification head removed. This model generates a sequence of feature maps $\{ \Phi_{l}(\mathcal{J}_{i}) \}_{i=1}^{N}$, corresponding to the $l^{\text{th}}$ layer of the AlexNet architecture. These feature maps are then used to compute the LPIPS score $\mathcal{S}_{lpips}$~\cite{zhang2018unreasonable}, for each pair $\{ \mathcal{J}_{i-1}, \mathcal{J}_{i} \}$, which assesses their perceptual similarity:
 \begin{equation} \label{lpipsEq}
     \mathcal{S}_{lpips} = \sum_{l} \lambda_{l} \left[ \sum_{l} \| \hat{\Phi}_{c}^{c}(\mathcal{J}_{i}) - \hat{\Phi}_{l}^{c}(\mathcal{J}_{i-1}) \|_{2}^{2} \right]
 \end{equation}
 Here, $\hat{\Phi}_{l}^{c}(\mathcal{J}{i})$ denotes the $c^{\text{th}}$ channel of the normalised feature map $\hat{\Phi}_{l}(\mathcal{J}_{i}) = \frac{\Phi_{l}(\mathcal{J}_{i})}{\| \Phi_{l}(\mathcal{J}_{i}) \|_{2}}$. A lower $\mathcal{S}_{\text{lpips}}$ score indicates greater perceptual similarity between $\mathcal{J}_{i-1}$ and $\mathcal{J}_{i}$, and thus higher confidence in the warped depth map $\mathcal{D}_{i}^{warp'}$. Conversely, a higher LPIPS score suggests reduced perceptual similarity, leading to increased reliance on the original depth estimate $\mathcal{D}_{i}^{mde}$. The final MAPIS-Depth maps are then computed through a weighted linear blending of $\mathcal{D}_{i}^{warp'}$ and $\mathcal{D}_{i}^{mde}$ as follows:
 \begin{equation}
 \hat{\mathcal{D}}_{i} = \mathcal{S}_{lpips} \cdot \mathcal{D}_{i}^{warp'} + (1 - \mathcal{S}_{lpips}) \cdot \mathcal{D}_{i}^{mde}
 \end{equation}
 Here, $\hat{\mathcal{D}}_{i}$ denotes the refined MAPIS-Depth map corresponding to the $i^{\text{th}}$ image in the sequence. The impact of this acceleration on performance and accuracy is further analysed in Section~\ref{experimental_discussion}.

 \subsection{WEMA-RTDL module}
 The refined MAPIS-Depth maps $\hat{\mathcal{D}}_{i}$ are combined with their corresponding RGB images $\mathcal{J}_{i}$ to form pseudo-RGBD frames, denoted as $\mathcal{F}_{i}^{rgbd}$. From the set $\{ \mathcal{F}_{i}^{rgbd} \}_{i=1}^{N}$, each frame $\mathcal{F}_{j}^{rgbd}$, starting from $\mathcal{F}_{0}^{rgbd}$, is selected as a keyframe $K_{j}^{ref}$, where $j = \alpha \cdot m$, with $\alpha = 2$ as a fixed non-negative integer and $m \in [0, \lfloor N/2 \rfloor] \cap \mathbb{Z}$. Each intermediate frame to be aligned with a reference keyframe is denoted as $K_{j,r}^{track}$, while the subsequent keyframe $K_{j+1}^{ref}$ is aligned with respect to the preceding one, $K_{j}^{ref}$.
 \par
 To enable robust frame alignment under varying motion conditions, multi-scale image pyramids are constructed for each frame pair. Specifically, pyramids $\{ K_{j,r}^{track,\Lambda} \}_{\Lambda=1}^{\Lambda_{pyr}}$ and $\{ K_{j}^{ref,\Lambda} \}_{\Lambda=1}^{\Lambda_{pyr}}$ are generated for $K_{j,r}^{track}$ and $K_{j}^{ref}$, respectively. Here, $\Lambda_{pyr} = 5$ denotes the total number of pyramid levels, with each level downsampled by a factor of $d = 0.5$. These pyramids facilitate multi-resolution analysis, allowing the alignment to begin at a coarse spatial scale—useful for handling large inter-frame displacements and progressively refining the transformation at higher resolutions. This hierarchical strategy ensures robust and precise alignment of pseudo-RGBD frames.

 \subsubsection{Rotation-Translation Decoupling}
 The estimation of endoscope motion begins by initialising the rotation parameters $\boldsymbol{x}_{j,r}^{so3} \in \mathcal{SO}(3)$ and the combined rotation–translation parameters $\boldsymbol{x}_{j,r}^{se3} \in \mathcal{SE}(3)$ as identity matrices for the first pyramid level ($\Lambda = 1$) corresponding to $K_{j,r}^{track,\Lambda}$. The photometric error is computed as follows~\cite{hartley2003multiple}:
 \begin{equation}
     e_{j,r}^{k} = \| \mathcal{J}_{j}(\boldsymbol{p}) -  \mathcal{J}_{j,r}(\boldsymbol{p^{'}})\|_{2}
 \end{equation}
 where $e_{j,r}^{k}$ denotes the photometric error at the $k^{\text{th}}$ optimisation iteration. The term $\mathcal{J}_{j}(\boldsymbol{p})$ represents the intensity at pixel $\boldsymbol{p}$ in the reference image $K{j}^{ref,\Lambda}$, and $\mathcal{J}_{j,r}(\boldsymbol{p}^{'})$ denotes the interpolated intensity at the transformed location $\boldsymbol{p}^{'}$ in the tracked frame $K_{j,r}^{track,\Lambda}$. The transformed position $\boldsymbol{p}^{'}$ is obtained via:
 \begin{equation*}
     \boldsymbol{p^{'}} = \mathcal{C}_{j} \mathcal{T}_{j}^{-1} \mathcal{T}_{j,r} \mathcal{C}_{j,r}^{-1} \mathcal{D}_{j,r} \boldsymbol{p}
 \end{equation*}
 where $\mathcal{C}_{j}$ and $\mathcal{C}_{j,r}$ are the intrinsic matrices corresponding to the reference and tracked frames, respectively, and $\mathcal{T}_{j}$ and $\mathcal{T}_{j,r}$ are the transformation matrices derived from $\boldsymbol{x}_{j}^{se3}$ and $\boldsymbol{x}_{j,r}^{se3}$. The matrix $\boldsymbol{x}_{j,r}^{se3}$ is constructed as:
 \begin{equation} \label{matrot}
     \boldsymbol{x}_{j,r}^{se3} = \begin{bmatrix}
         \boldsymbol{x}_{j,r}^{so3} \odot \boldsymbol{x}_{j, rot}^{se3} & \boldsymbol{x}_{j,r}^{so3} \odot \boldsymbol{x}_{j, trans}^{se3} \\
         \boldsymbol{0} & 1
     \end{bmatrix}
 \end{equation}
 Here, $\boldsymbol{x}_{j, rot}^{se3}$ and $\boldsymbol{x}_{j, trans}^{se3}$ represent the $3 \times 3$ rotation and $3 \times 1$ translation components, respectively, and $\odot$ denotes standard matrix multiplication.

 \subsubsection{Optimisation}
 To optimise the rotation parameters $\boldsymbol{x}_{j,r}^{so3}$, a dog-leg optimisation scheme is adopted~\cite{makaev2019software}, initialised with a trust region radius $\rho \in \mathbb{R}^{+}$. This radius governs the step size during each iteration by balancing between the steepest descent and Gauss–Newton directions, thereby facilitating efficient convergence~\cite{gauss1809theoria, curry1944method}:
 \begin{align}
     & \rho_{gn} = -\left( \boldsymbol{J}^{T} \boldsymbol{J} \right)^{-1} \boldsymbol{J}^{T} e_{j,r}^{k}\\
     & \rho_{sd} = - \boldsymbol{J}^{T} e_{j,r}^{k}
 \end{align}
 Here, $\rho_{gn}$ and $\rho_{sd}$ denote the candidate step sizes $\Delta \boldsymbol{x}$ derived from the Gauss–Newton and steepest descent strategies, respectively, and $\boldsymbol{J}$ is the Jacobian matrix of partial derivatives of $e_{j,r}^{k}$ with respect to $\boldsymbol{x}_{j,r}^{se3}$. The final update step $\Delta \boldsymbol{x}$ is selected according to the following rule~\cite{makaev2019software}:
 \begin{equation}
     \Delta \boldsymbol{x} = \begin{cases}
         \rho_{gn} & \mbox{ if } \| \rho_{gn}\| \leq \rho \\
         \rho \frac{\rho_{sd}}{\| \rho_{sd} \|} & \mbox{ if } \| \rho_{gn}\| > \rho \mbox{ and } \| \rho_{sd} \| > \rho \\
         (1 - s)h \rho_{sd} + s\rho_{gn} & \mbox{ if } \| \rho_{gn}\| > \rho \mbox{ and } \| \rho_{sd} \| < \rho \\
         
     \end{cases}
 \end{equation}
 where $h = \frac{\| \rho_{sd} \|^{2}}{\| \boldsymbol{J} \rho_{sd} \|^{2}}$ and $s$ is selected such that $\| \rho_{gn} \| = \| \rho_{sd} \| = \rho$. The trust region radius is adaptively adjusted, effectively navigating the optimisation process and providing an estimate for rotation parameters as follows:
 \begin{equation} \label{elfeq}
     _{est}\boldsymbol{x}_{j,r}^{so3} = \boldsymbol{x}_{j,r}^{so3} + \Delta\boldsymbol{x}_{rot}
 \end{equation}
 where $\Delta \boldsymbol{x}_{rot}$ denotes the rotational component of the step $\Delta \boldsymbol{x}$.
 \par
 Following rotation optimisation at the coarsest pyramid level ($\Lambda = 1$), the full transformation parameters $\boldsymbol{x}_{se3} \in \mathcal{SE}(3)$, encompassing both rotation and translation, are optimised for levels $\Lambda \geq 2$. For $\Lambda = 2$, the previously estimated rotation $_{est}\boldsymbol{x}_{j,r}^{so3}$ is used to initialise the rotational component of $\boldsymbol{x}_{se3}$, yielding an initial estimate denoted as $_{in}\boldsymbol{x}_{j,r}^{se3}$. This initial transformation is then refined using the same optimisation procedure described for $_{est}\boldsymbol{x}_{j,r}^{so3}$, with the exception of the update rule in Equation~\eqref{matrot}. Since both rotation and translation are now subject to refinement, the update follows:
 \begin{equation}
     \boldsymbol{x}_{j,r}^{se3} = \mbox{\ }_{in}\boldsymbol{x}_{j,r}^{se3} \cdot \boldsymbol{x}_{j}^{se3}
 \end{equation}
 where $\cdot$ denotes standard matrix multiplication. After performing dog-leg optimisation, the refined pose is computed as:
 \begin{equation}
     _{est}\boldsymbol{x}_{j,r}^{se3} \leftarrow \boldsymbol{x}_{j,r}^{se3} + \Delta \boldsymbol{x}
 \end{equation}
  This iterative process provides robust endoscope pose estimation in terms of $\boldsymbol{x}_{se3}$ and ensures accurate registration of $\mathcal{F}_{i}^{rgbd}$. The estimated $\mathcal{SE}(3)$ pose $\boldsymbol{x}_{se3}$ is then used to update the camera pose for the tracked frame $K_{j,r}^{track}$.

  \subsubsection{Regularisation}
  To stabilise pose estimates and minimise jitter in $\mathcal{F}_{i}^{rgbd}$ registration, the Exponential Moving Average (EMA) technique is applied to the estimated poses. Given the current estimated pose $\boldsymbol{x}_{se3}^{i}$ and the two previous poses $\boldsymbol{x}_{se3}^{i-1}$ and $\boldsymbol{x}_{se3}^{i-2}$, the regularised pose $\boldsymbol{x}_{se3}^{i,reg}$ is computed as:
 \begin{equation}
     \boldsymbol{x}_{se3}^{i,reg} = (\alpha \boldsymbol{x}_{se3}^{i} + \beta \boldsymbol{x}_{se3}^{i-1} + \gamma \boldsymbol{x}_{se3}^{i-2}) s + \boldsymbol{x}_{se3}^{i} (1-s)
 \end{equation}
 Here, $\alpha$, $\beta$, and $\gamma$ are weighting factors satisfying $\alpha + \beta + \gamma = 1$, and $s = 1 - \exp(-i*\omega)$ adjusts for the accumulated frame count $i$ using the scaling factor $\omega \in \mathbb{R}^{+}$. The coefficient $s$ ensures that the regularisation becomes more effective as the frame number becomes larger, since at large frame numbers drift becomes more influencing and hence regularisation stabilises the endoscope poses. This approach ensures smoother camera pose transitions by incorporating historical pose data, effectively reducing noise and enhancing registration robustness in the MMIS environment.

 \subsection{$3D$ Reconstruction}
 This module integrates the registered pseudo-RGBD frames using a TSDF approach~\cite{recasens2021endo}, which efficiently fuses per-frame depth and pose information into a coherent \textit{3D} surface representation. Depth measurements are accumulated within a voxel grid as signed distances relative to the nearest observed surface, and are iteratively refined to suppress noise and artefacts~\cite{perera2015motion}. Finally, the Marching Cubes algorithm~\cite{lorensen1998marching} is applied to extract a smooth and topologically consistent mesh, thereby enhancing monocular imagery with geometrically accurate tissue surface reconstructions.

   \section{Experimental Results and Discussion}

 \subsection{Dataset} \label{datasets}
 The proposed framework is rigorously evaluated on two benchmark datasets: HEVD~\cite{recasens2021endo} and SCARED~\cite{allan2021stereo}. The HEVD dataset consists of monocular endoscopic video sequences that are representative of clinical scenarios, including motion blur, specular highlights, physiological tissue deformation, and smoke occlusion, thereby serving as a representative testbed for reconstruction under surgically realistic conditions. In contrast, the SCARED dataset provides precisely calibrated ground truth information derived from structured light projection on porcine anatomy. Its millimetre-scale accuracy and controlled setup enable quantitative assessment in the presence of anatomical complexity and occlusions.

 \subsection{Evaluation Metrics} \label{metricSection}
 To assess the performance of the proposed framework, the consistency in depth estimation, accuracy of endoscope's estimated pose, and scale-awareness are evaluated. 
 \par
 Depth consistency is evaluated through an ORB-based point-to-point correspondence analysis across temporally adjacent frames~\cite{rublee2011orb}. For each pair of consecutive point clouds reconstructed from predicted depth maps, matched 3D points are sampled and the displacement  between them are computed. Bar plots of these displacements, along with mean and median statistics, provide a quantitative indicator of temporal smoothness and structural coherence in the estimated depth. Reduced peak displacements signal higher geometric consistency across frames.
 \par
 Pose estimation results are grouped into four categories: (i) per-axis errors, comprising mean translation ($T_{\mathrm{avg}}$) and mean rotation error ($R_{\mathrm{avg}}$), which quantify axis-wise pose drift across the trajectory; (ii) global errors, which include root mean square error (RMSE), terminal translation error ($T_{\mathrm{fin}}$), and terminal rotation error ($R_{\mathrm{fin}}$), reflecting the accumulated deviation at the end of the sequence; (iii) frame-level errors, measured via relative pose error (RPE), as well as maximum translation ($T_{\mathrm{max}}$) and rotation errors ($R_{\mathrm{max}}$), to capture transient spikes and inter-frame drift; and (iv) statistical proportions, reporting the percentage of poses with error below the mean ($T < T_{\mathrm{avg}}$ and $R < R_{\mathrm{avg}}$), which provide a measure of tracking robustness by indicating the frequency of low-error pose estimates along the trajectory~\cite{sturm2012benchmark, luperto2021predicting}.
 \par
 Scale awareness is assessed using the Trajectory Length Ratio (TLR), defined as the ratio between the estimated trajectory length and the ground truth trajectory length. A TLR close to 1 indicates consistent scaling in estimated poses, while significant deviations from unity indicate scale drift. This metric thus serves as a proxy for global scale fidelity, particularly important in monocular settings where scale ambiguity is a critical challenge.

 \subsection{Results and Discussion}\label{experimental_discussion}
  All experiments are conducted on an NVIDIA GeForce RTX 4080 GPU~\cite{wagner2025influence}, leveraging its high computational throughput for efficient execution of the framework. The experimental protocol is structured to comprehensively evaluate the proposed framework across diverse MMIS scenarios, including challenging intraoperative conditions such as illumination variation, occlusion, deformation, and smoke. The results encompass both qualitative visualisations and quantitative metrics, spanning depth consistency, pose regularity, and scale fidelity.
  \par
\begin{figure}
    \centering
    \includegraphics[width=1.0\linewidth]{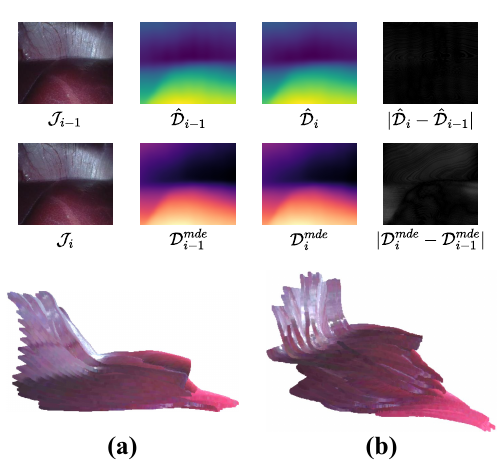}
    \caption{Temporal consistency assessment with MAPIS-Depth and MDE~\cite{bochkovskii2024depth}. Top: consecutive frames $\mathcal{J}_{i-1}$ and $\mathcal{J}_i$ with their respective depth predictions and absolute difference maps. Bottom: registered pseudo-RGBD reconstructions show improved geometric continuity using MAPIS-Depth (a) compared to MDE (b). Visualised on the HEVD dataset~\cite{recasens2021endo}.}
    \label{registration}
 \end{figure}

  The first experiment, illustrated in Fig.~\ref{registration}, evaluates the temporal consistency of depth estimation by the proposed MAPIS-Depth module on an HEVD dataset~\cite{recasens2021endo}. Each row in the upper panel of Fig.~\ref{registration} presents a pair of temporally adjacent frames, $\mathcal{J}_{i-1}$ and $\mathcal{J}_{i}$, along with their corresponding depth maps. The top row visualises the depth predictions $\hat{\mathcal{D}}_{i-1}$ and $\hat{\mathcal{D}}_i$ from MAPIS-Depth, while the bottom row shows depth maps $\mathcal{D}^{mde}_{i-1}$ and $\mathcal{D}^{mde}_i$~\cite{bochkovskii2024depth}. The final column in both rows presents the absolute difference maps: $|\hat{\mathcal{D}}_i - \hat{\mathcal{D}}_{i-1}|$ and $|\mathcal{D}^{mde}_i - \mathcal{D}^{mde}_{i-1}|$, respectively. It is evident that the MAPIS-Depth predictions yield significantly lower inter-frame residuals, indicating smoother and temporally coherent depth transitions. In contrast, the MDE predictions show elevated residuals and inconsistent geometric profiles across adjacent frames, highlighting the lack of temporal regularity.
  \par
\begin{figure}
    \centering
    \begin{tabular}{cc}
    \includegraphics[width=0.45\linewidth]{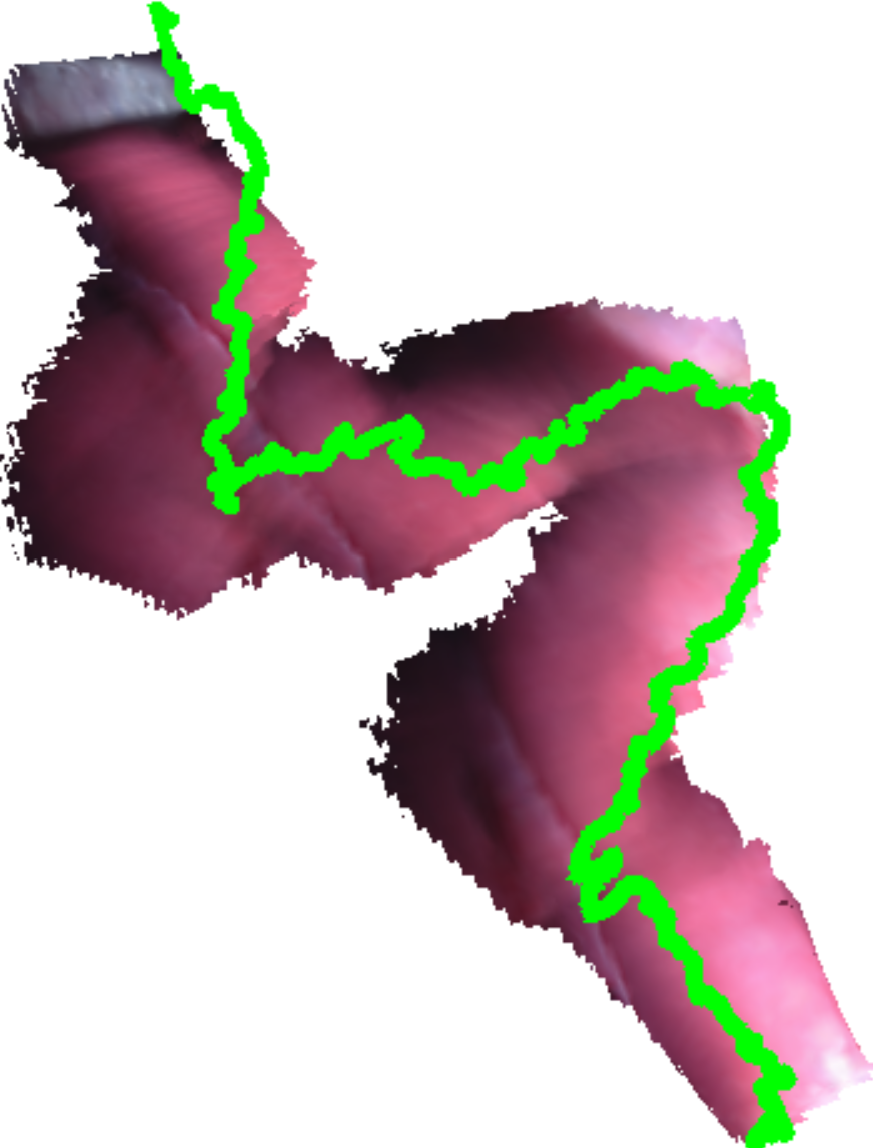} & \includegraphics[width=0.5\linewidth]{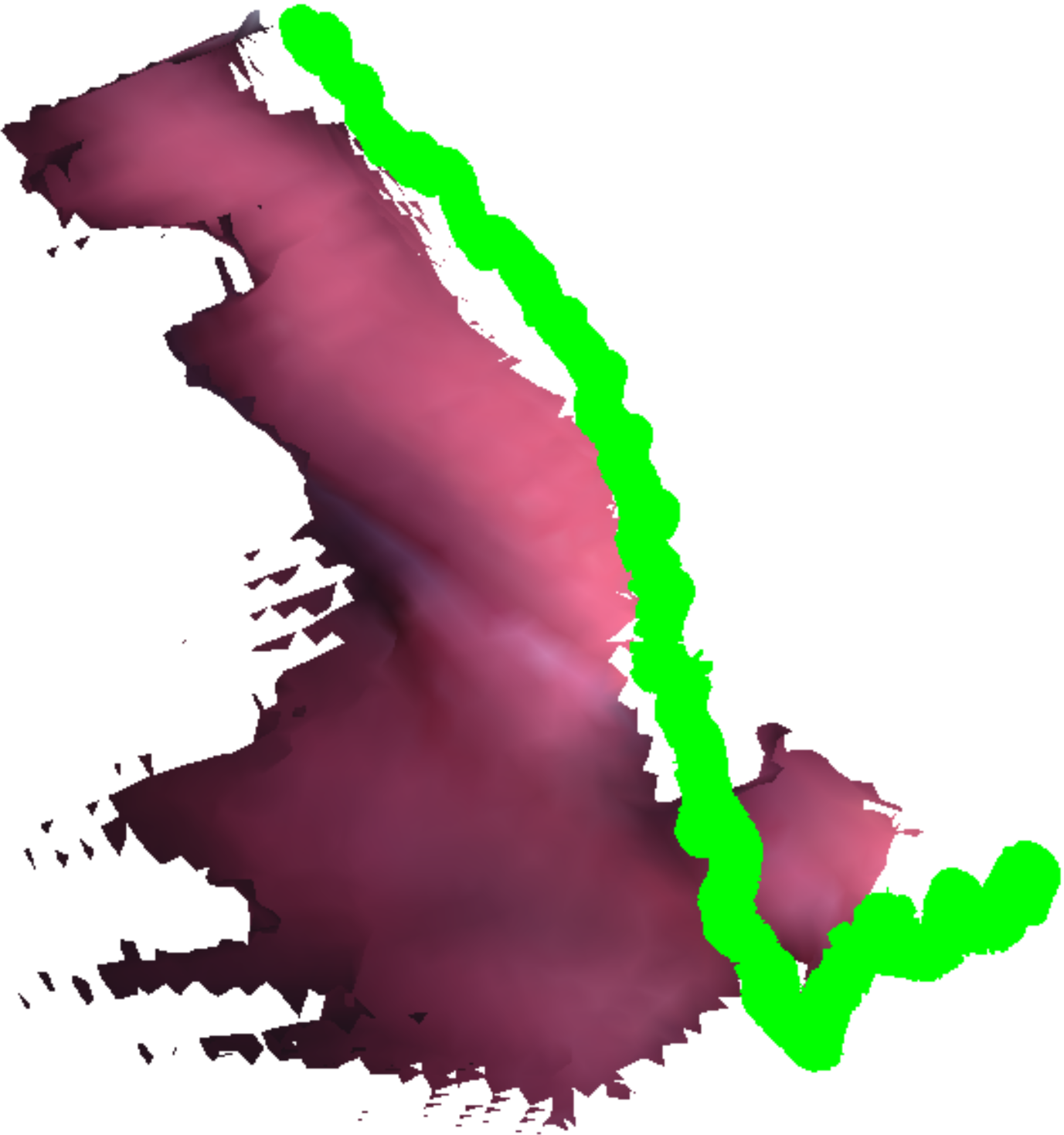}\\
    \textbf{(a)} & \textbf{(b)}
    \end{tabular}
    \caption{Reconstructed surfaces with estimated camera trajectories (green) on an HEVD dataset~\cite{recasens2021endo}: (a) MAPIS-Depth shows smoother paths and denser surfaces; (b) MDE~\cite{bochkovskii2024depth} exhibits fragmentation and drift.}

    \label{registration_00}
 \end{figure}

  Further evidence is provided by the registered point clouds (RPCs) shown at the bottom of Fig.~\ref{registration}. Subfigure~(a), generated from MAPIS-Depth maps, demonstrates structurally coherent and artefact-free registration with smooth surface continuity across frames. In contrast, subfigure~(b), based on MDE maps, exhibits prominent discontinuities and misalignments, particularly at tissue folds and borders. The complete reconstructions corresponding to Fig.~\ref{registration}~(a) and \ref{registration}~(b) are demonstrated in Fig.~\ref{registration_00}~(a) and \ref{registration_00}~(b), respectively. These observations confirm that the temporal regularisation introduced by MAPIS-Depth contributes to more consistent depth estimates across time, which is vital for monocular \textit{3D} reconstruction in dynamic and low-texture surgical scenes.

  \begin{figure}[ht!]
     \centering
     \begin{tabular}{ccccccc}
     \multicolumn{7}{c}{\textbf{(a)}} \\
      \includegraphics[width=1.95cm, height=1.95cm]{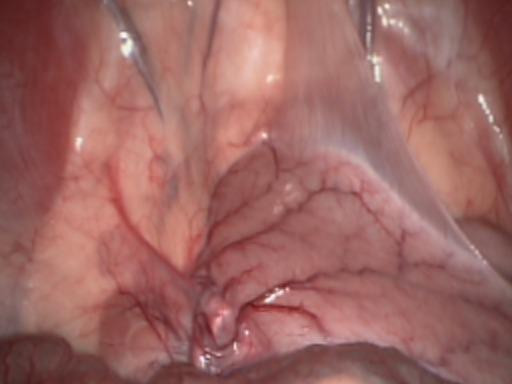} & \includegraphics[width=1.95cm, height=1.95cm]{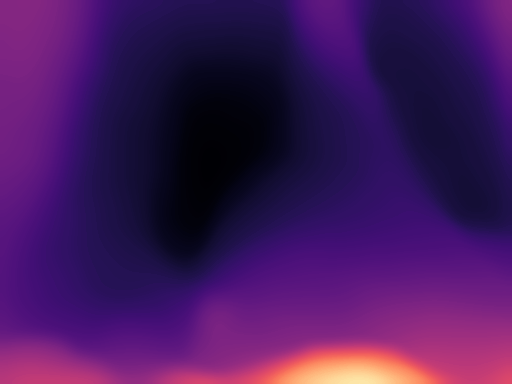} & \includegraphics[width=1.95cm, height=1.95cm]{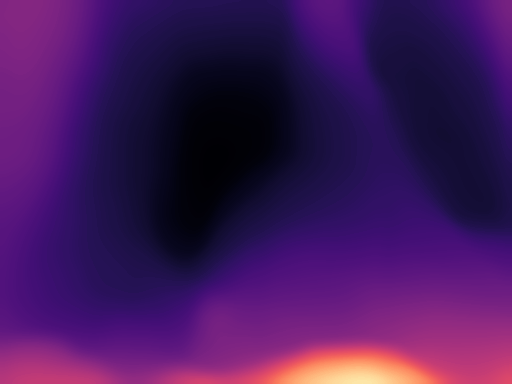} & \includegraphics[width=1.95cm, height=1.95cm]{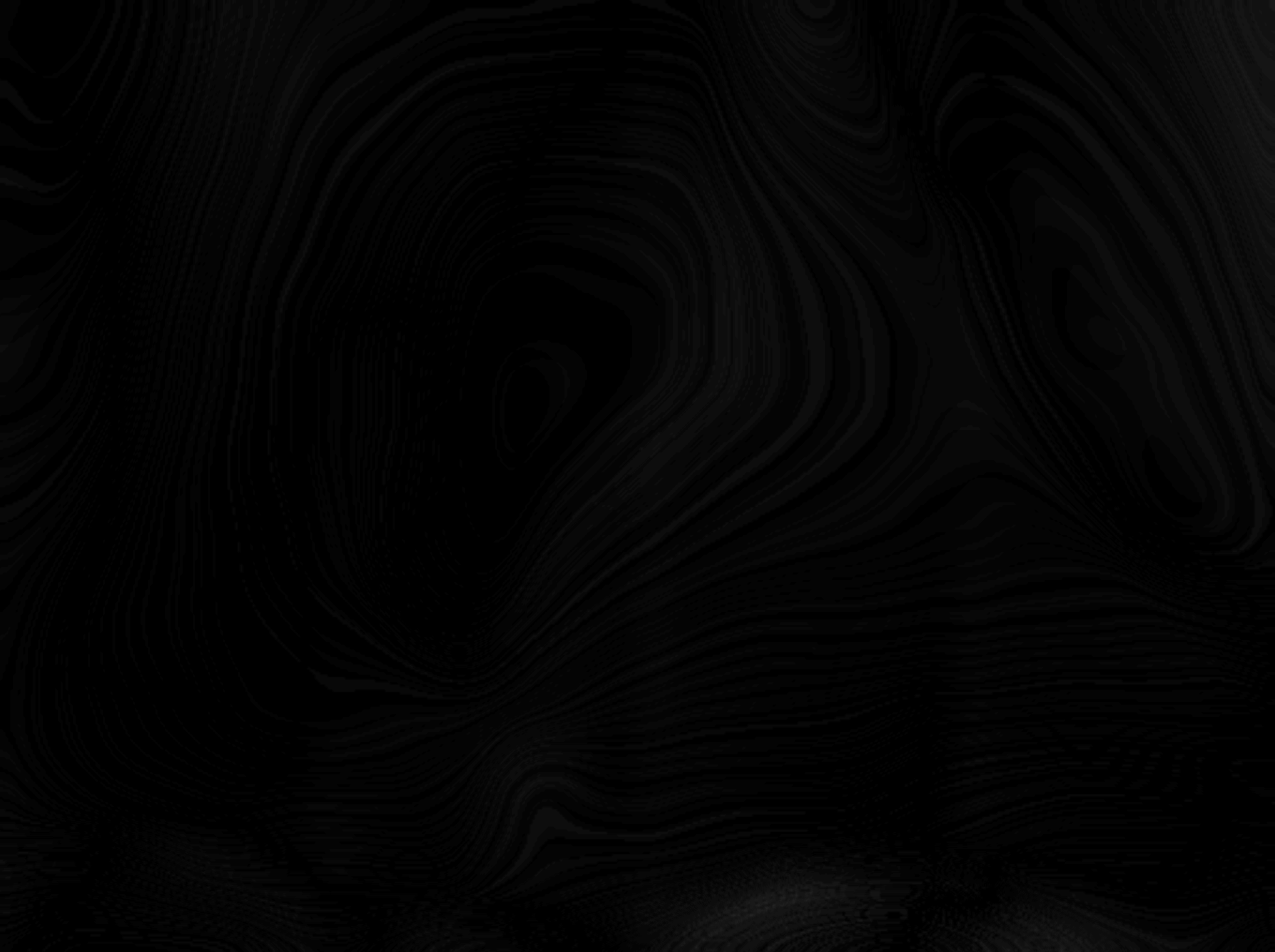} & \includegraphics[width=1.95cm, height=1.95cm]{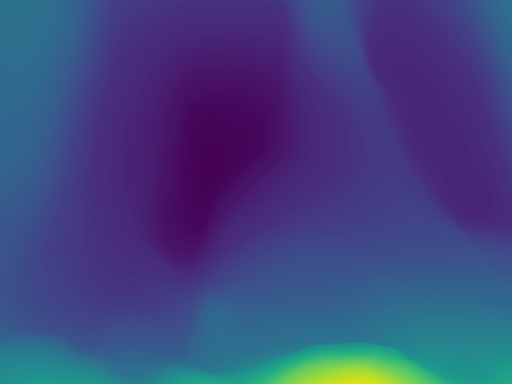} & \includegraphics[width=1.95cm, height=1.95cm]{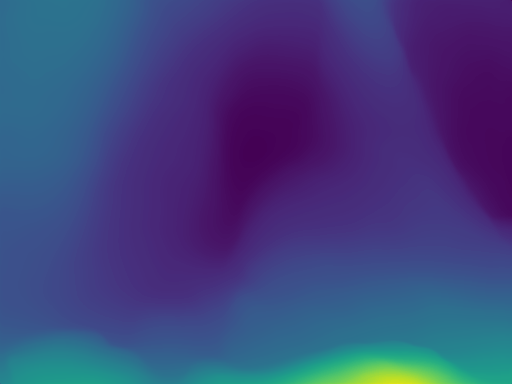} & \includegraphics[width=1.95cm, height=1.95cm]{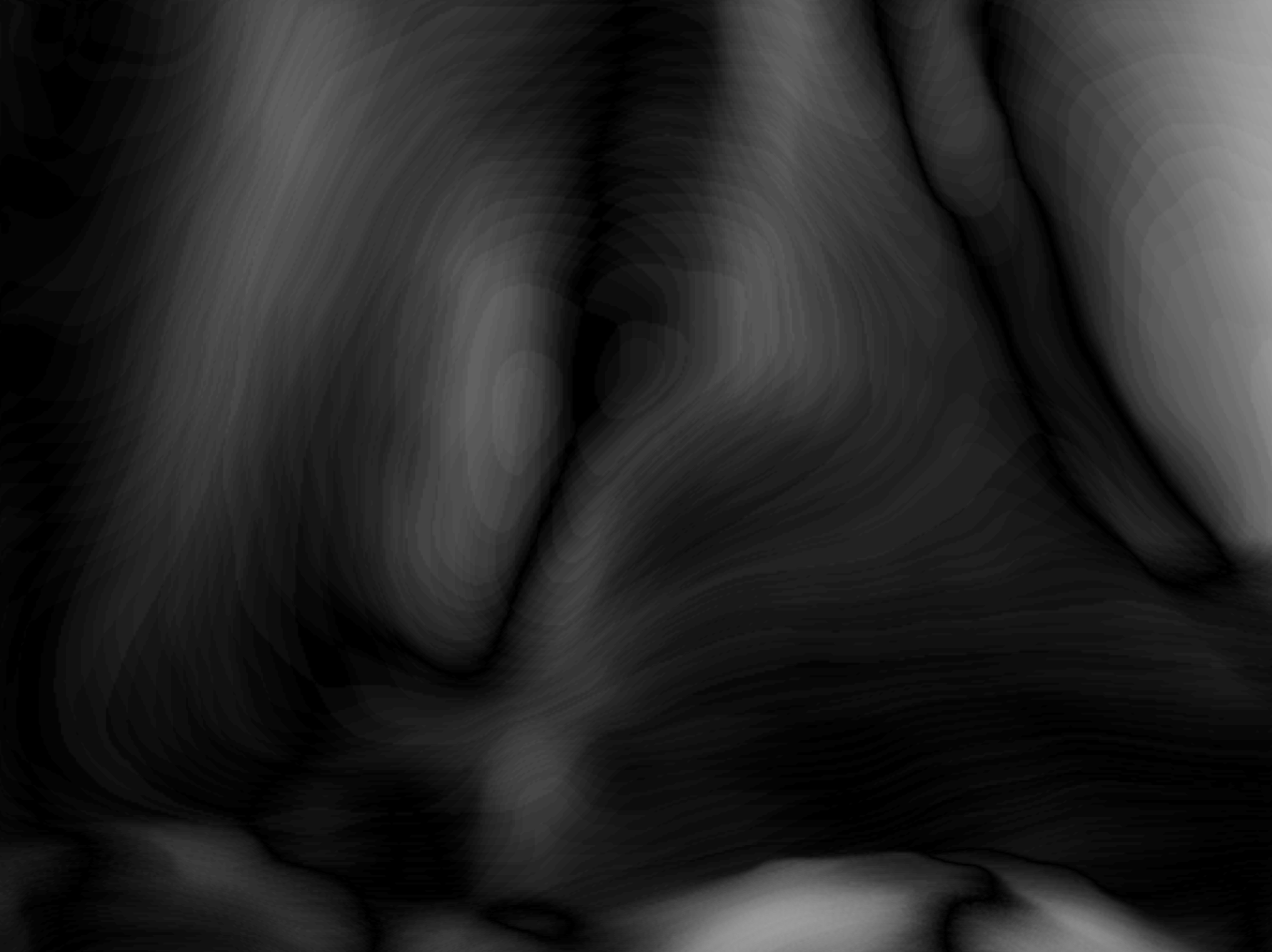}\\
      \multicolumn{7}{c}{\textbf{(b)}} \\
      \includegraphics[width=1.95cm, height=1.95cm]{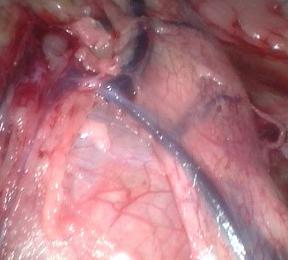} & \includegraphics[width=1.95cm, height=1.95cm]{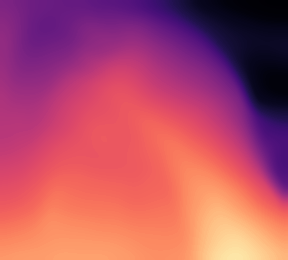} & \includegraphics[width=1.95cm, height=1.95cm]{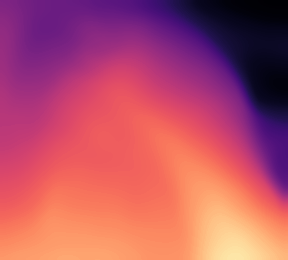} & \includegraphics[width=1.95cm, height=1.95cm]{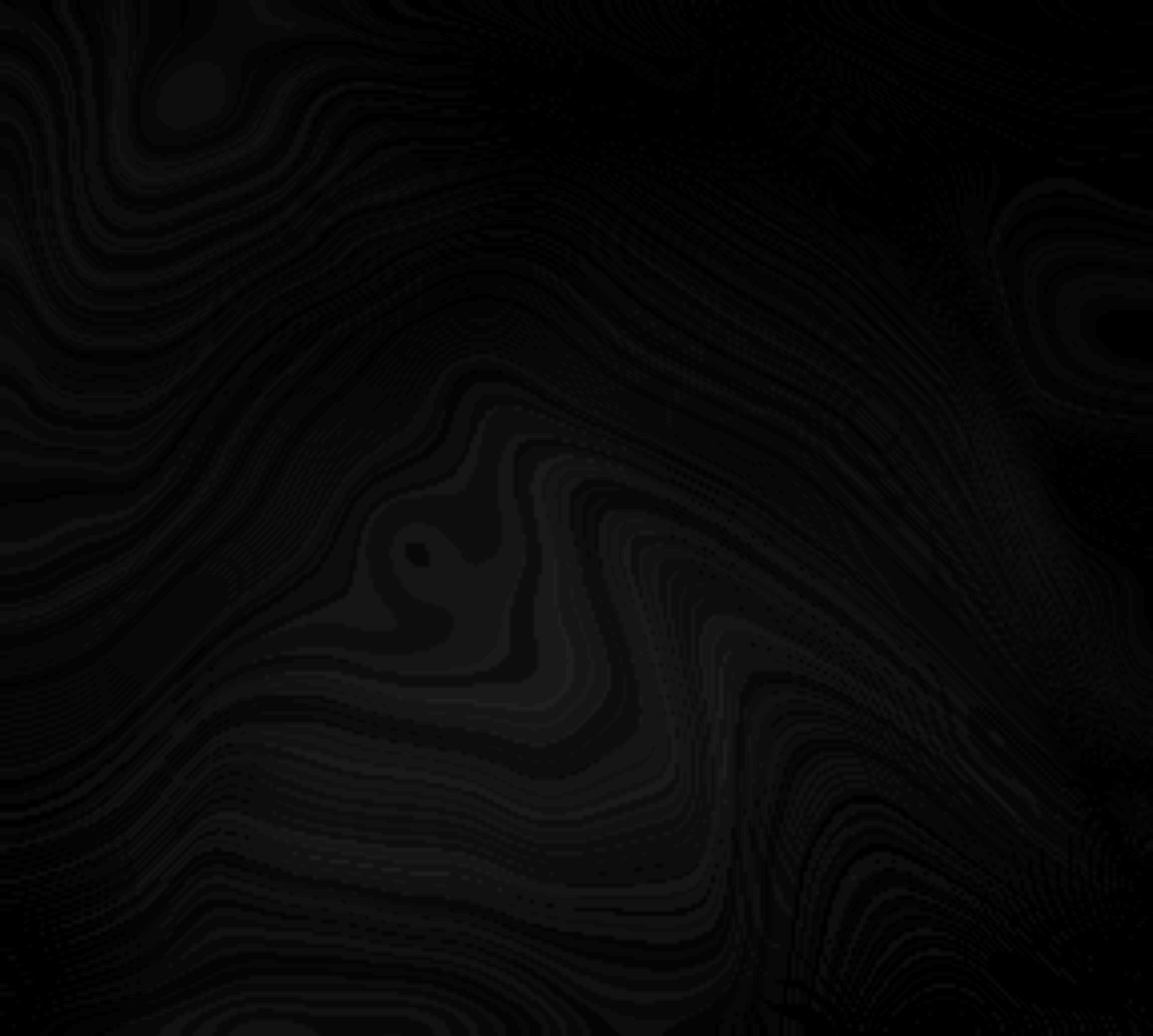} & \includegraphics[width=1.95cm, height=1.95cm] {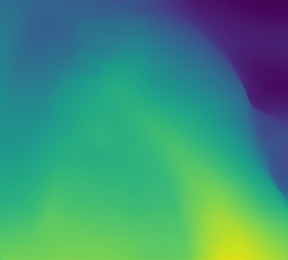} & \includegraphics[width=1.95cm, height=1.95cm]{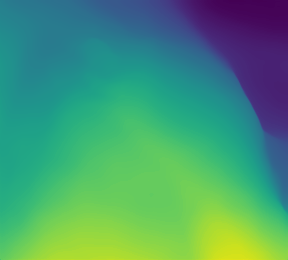} & \includegraphics[width=1.95cm, height=1.95cm]{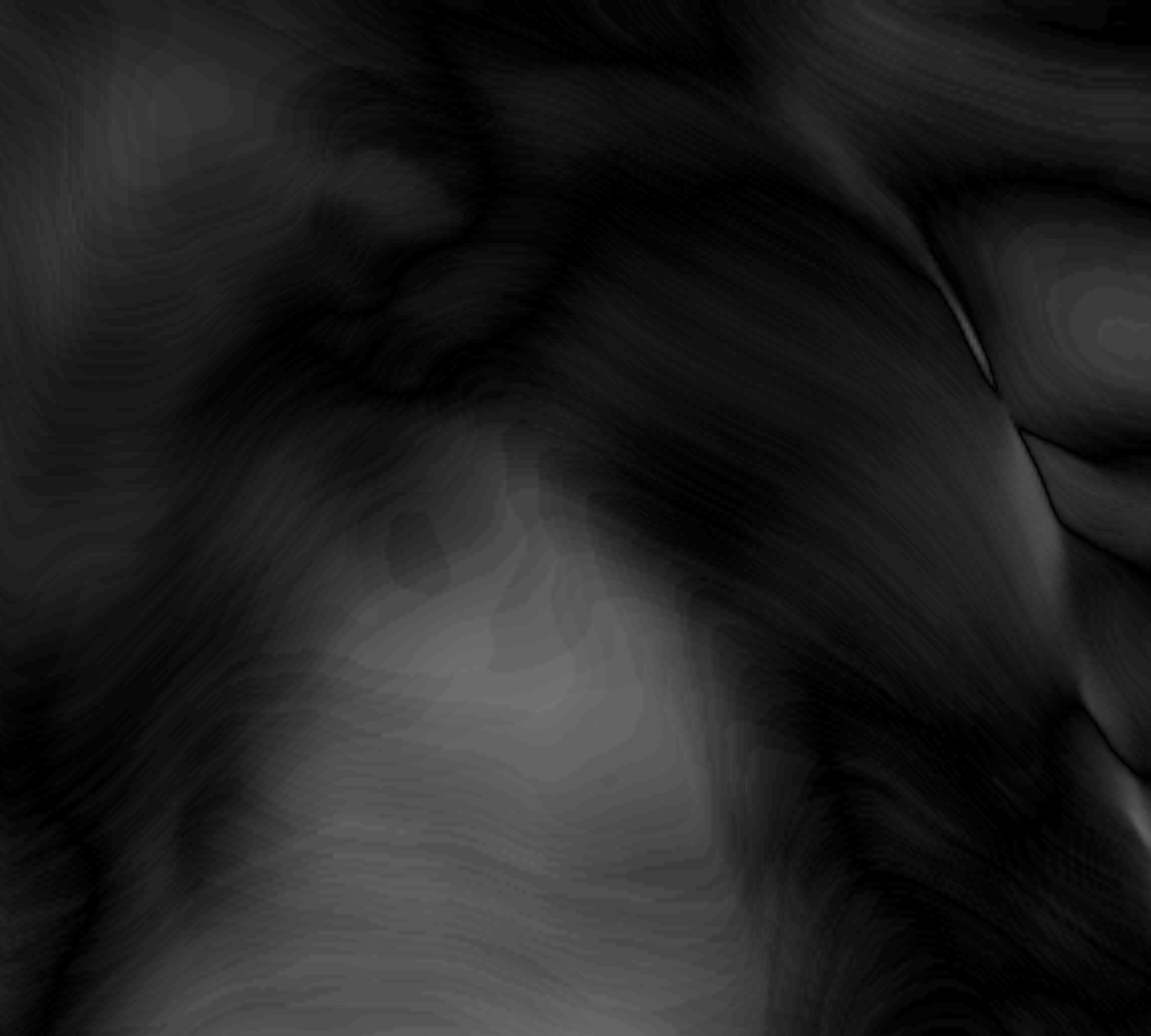}\\
      \multicolumn{7}{c}{\textbf{(c)}} \\
      \includegraphics[width=1.95cm, height=1.95cm]{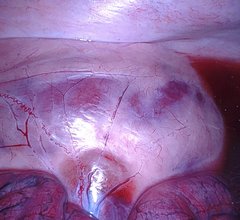} & \includegraphics[width=1.95cm, height=1.95cm]{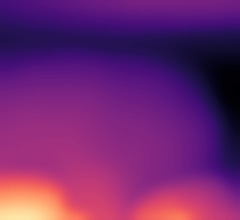} & \includegraphics[width=1.95cm, height=1.95cm]{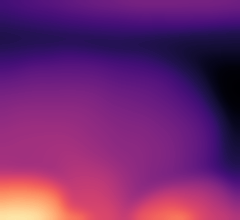} & \includegraphics[width=1.95cm, height=1.95cm]{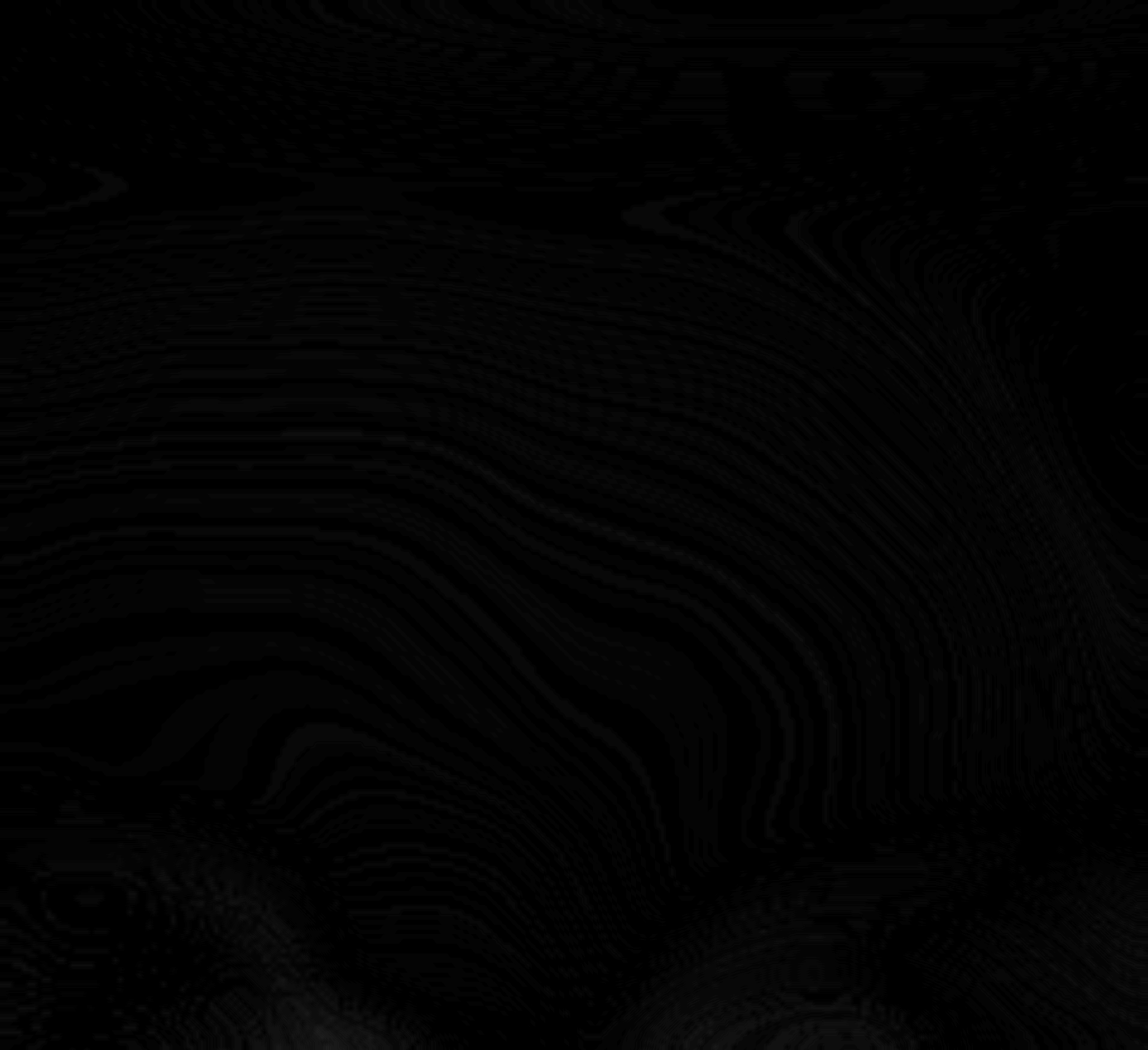} & \includegraphics[width=1.95cm, height=1.95cm]{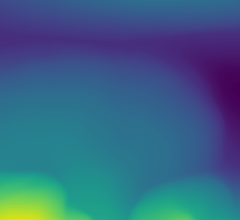} & \includegraphics[width=1.95cm, height=1.95cm]{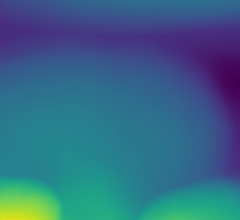} & \includegraphics[width=1.95cm, height=1.95cm]{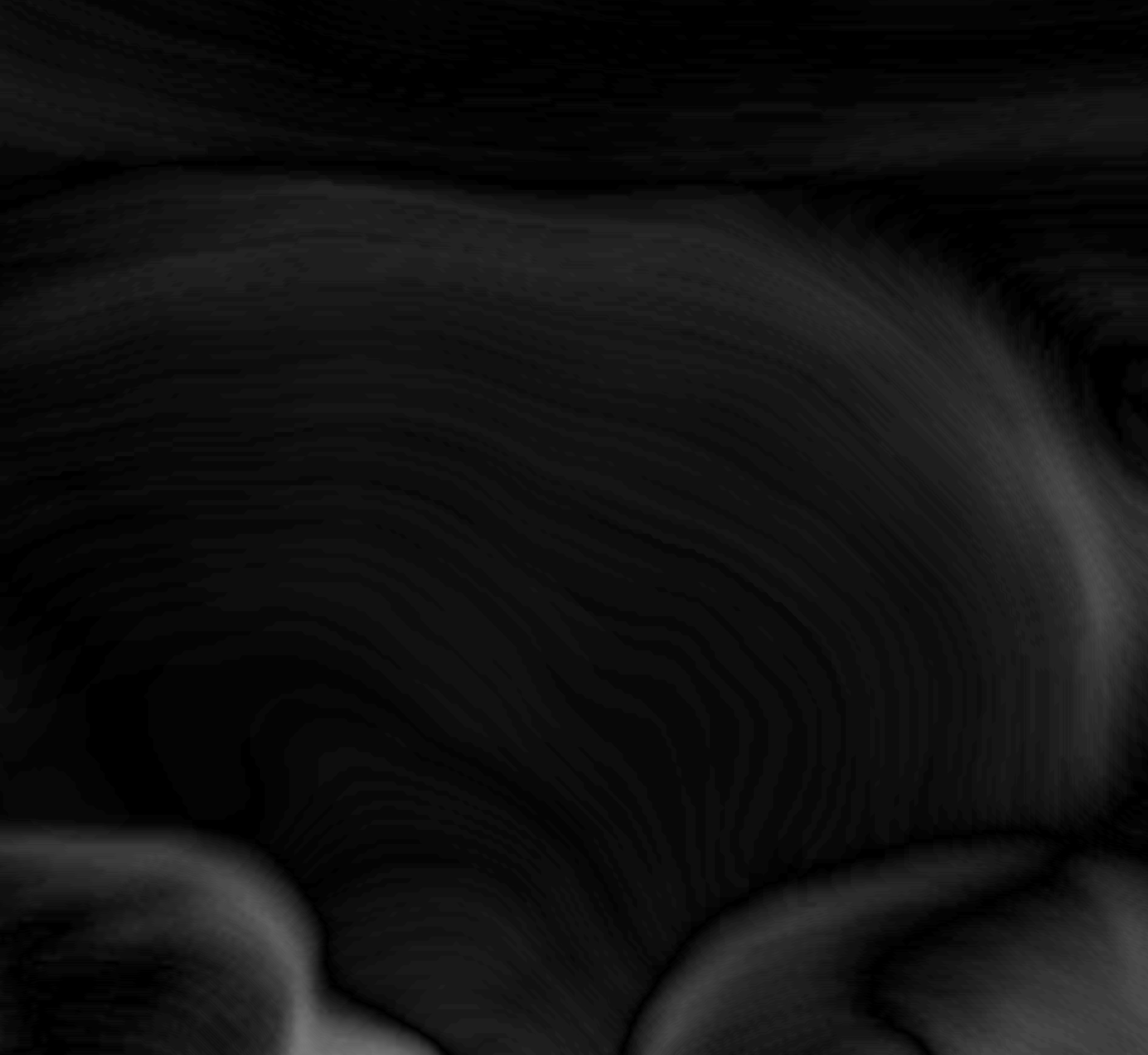}\\
      \multicolumn{7}{c}{\textbf{(d)}} \\
      \includegraphics[width=1.95cm, height=1.95cm]{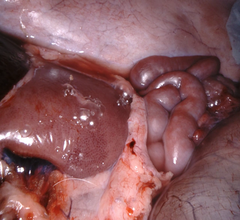} & \includegraphics[width=1.95cm, height=1.95cm]{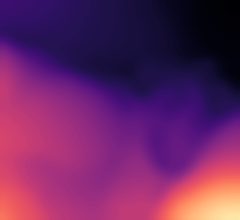} & \includegraphics[width=1.95cm, height=1.95cm]{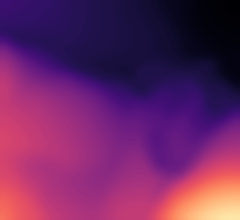} & \includegraphics[width=1.95cm, height=1.95cm]{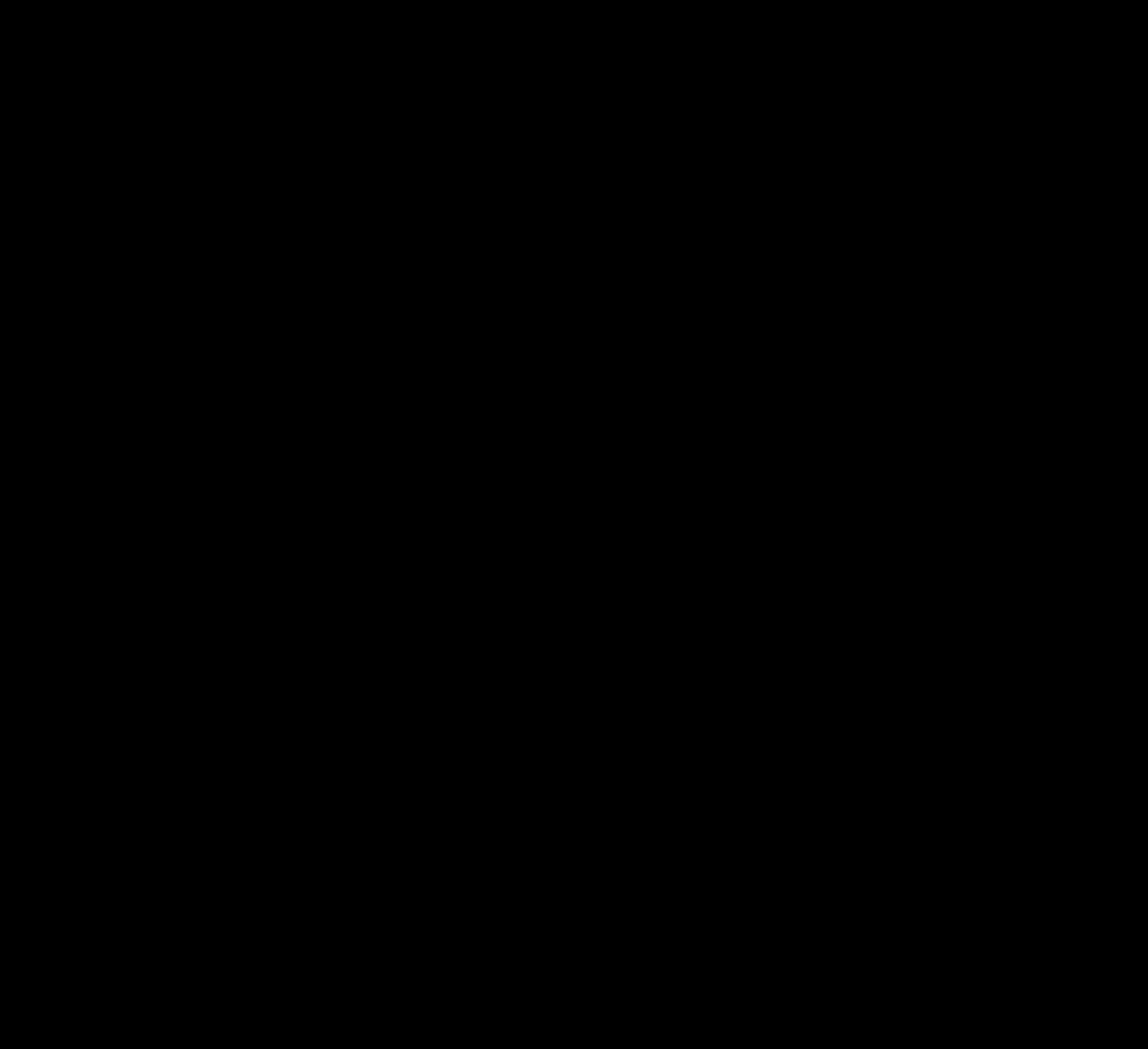} & \includegraphics[width=1.95cm, height=1.95cm]{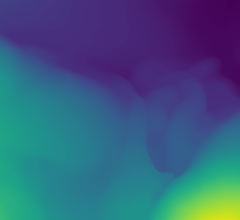} & \includegraphics[width=1.95cm, height=1.95cm]{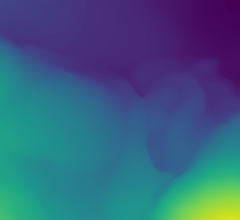} & \includegraphics[width=1.95cm, height=1.95cm]{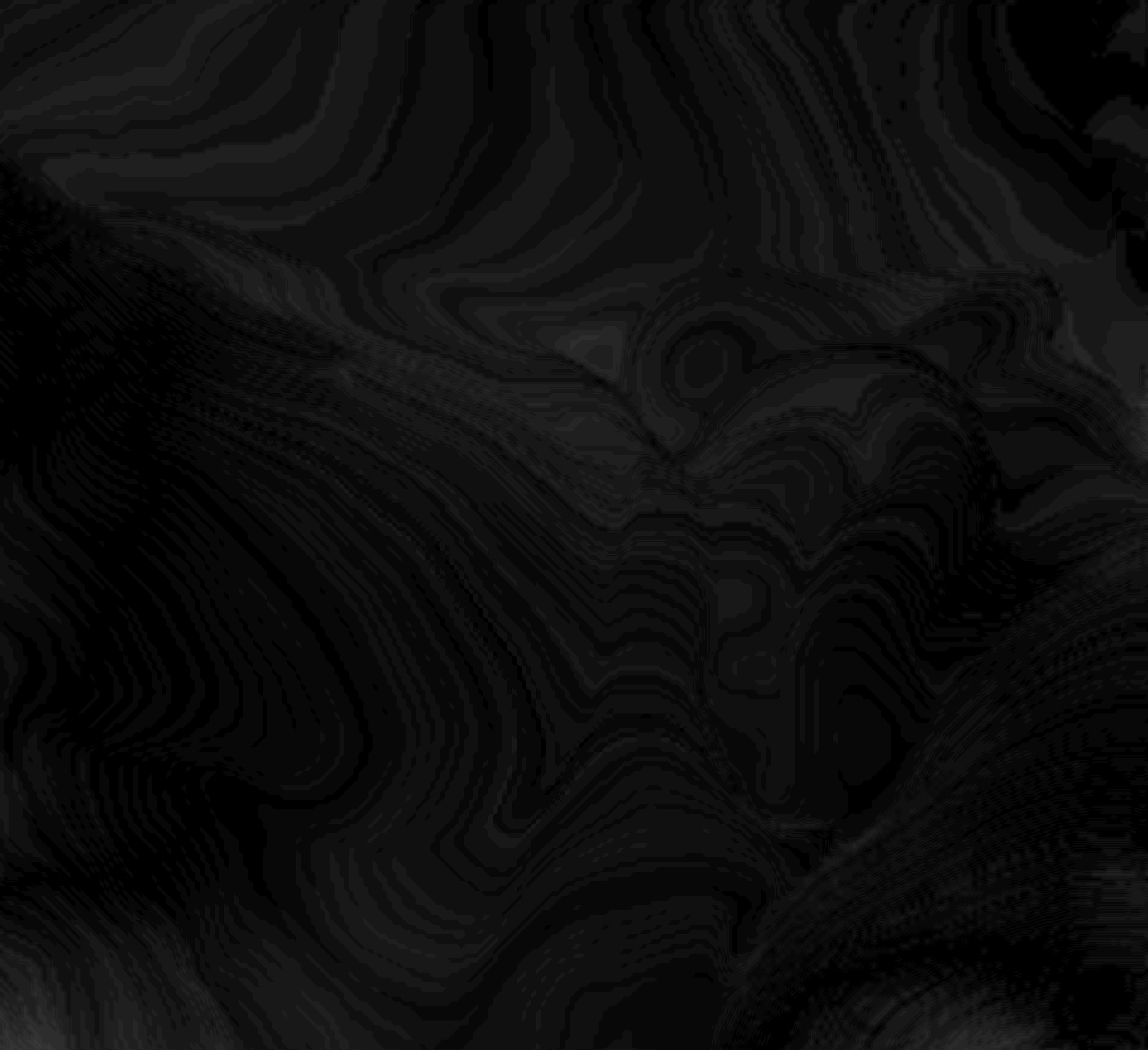} 
      \end{tabular}
      \caption{Temporal consistency evaluation across Various HEVD and SCARED datasets~\cite{recasens2021endo, allan2021stereo}. Representative images from the datasets (first column), MAPIS-Depth maps for two consecutive frames (second and third columns) and their pixel-wise differences (fourth column), MDEs~\cite{bochkovskii2024depth} (fifth and sixth columns) and their pixel-wise differences (seventh column).      \label{depthPIS0}}
 \end{figure}
 
  \par
 The second experiment extends the temporal consistency evaluation of the proposed MAPIS-Depth module across diverse HEVD and SCARED datasets~\cite{recasens2021endo, allan2021stereo}, as illustrated in Fig.~\ref{depthPIS0}. These datasets encompass varying surgical environments with differences in tissue structure, illumination, and camera motion patterns. Each row in Fig.~\ref{depthPIS0} represents one surgical scene. The first column shows representative RGB frames, followed by MAPIS-Depth predictions for two consecutive frames in the second and third columns. The fourth column presents the absolute difference between these MAPIS-Depth predictions, highlighting minimal inter-frame residuals and confirming strong temporal coherence. In contrast, the fifth and sixth columns show the corresponding MDEs~\cite{bochkovskii2024depth}, with their respective pixel-wise difference maps shown in the seventh column. These exhibit substantially higher residuals and geometric instability across frames.\\

 \par
\begin{figure*}[ht!]
     \centering
     \begin{tabular}{cccc}
     \textbf{(a)} & \textbf{(b)} & \textbf{(c)} & \textbf{(d)} \\
     \includegraphics[width=3.7cm, height=2.3cm]{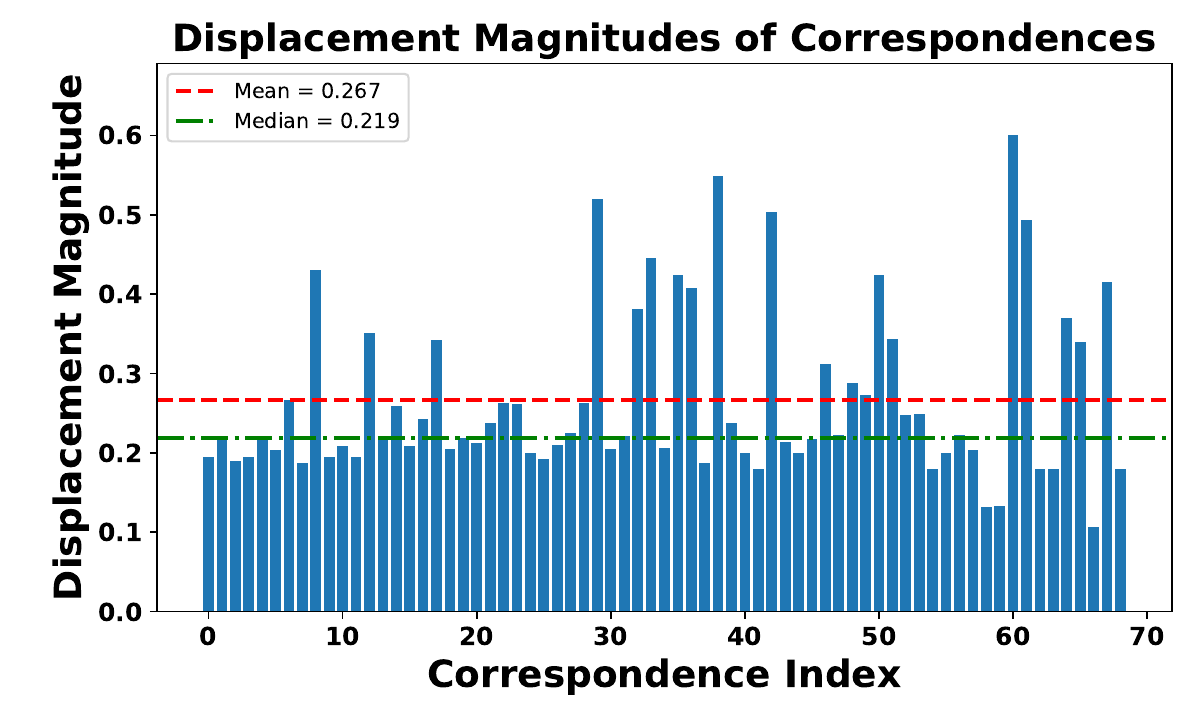} & \includegraphics[width=3.7cm, height=2.3cm]{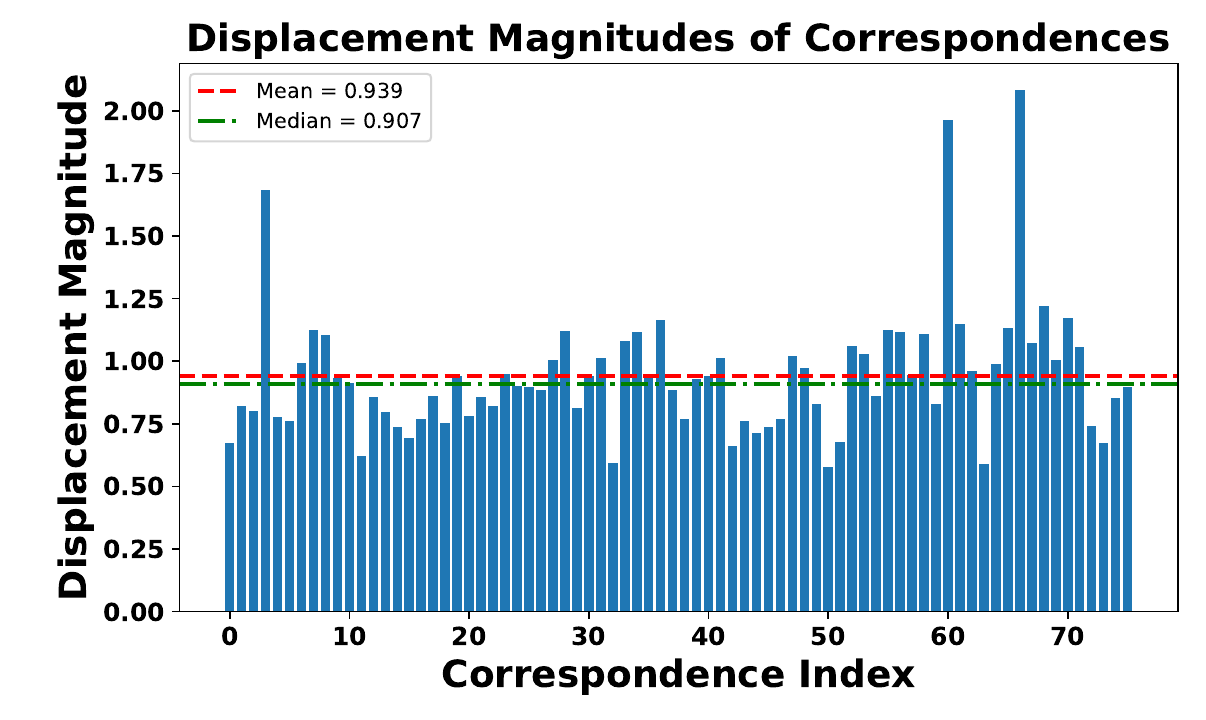} & \includegraphics[width=3.7cm, height=2.3cm]{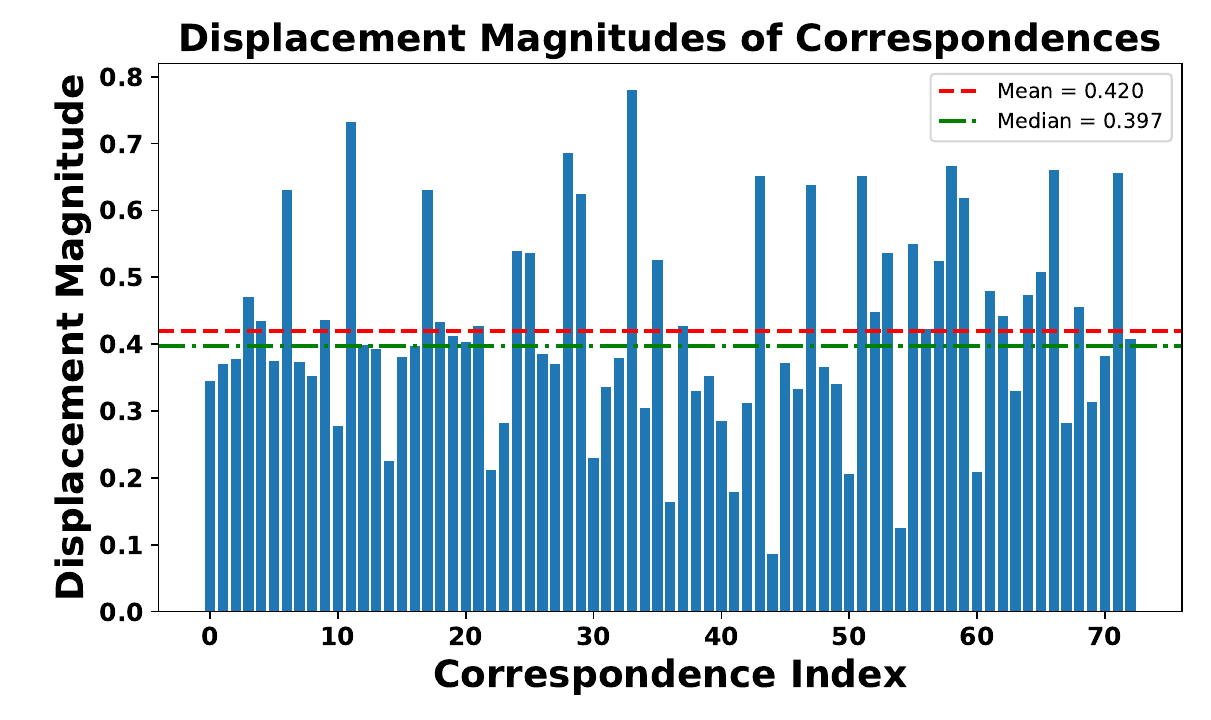} & \includegraphics[width=3.7cm, height=2.3cm]{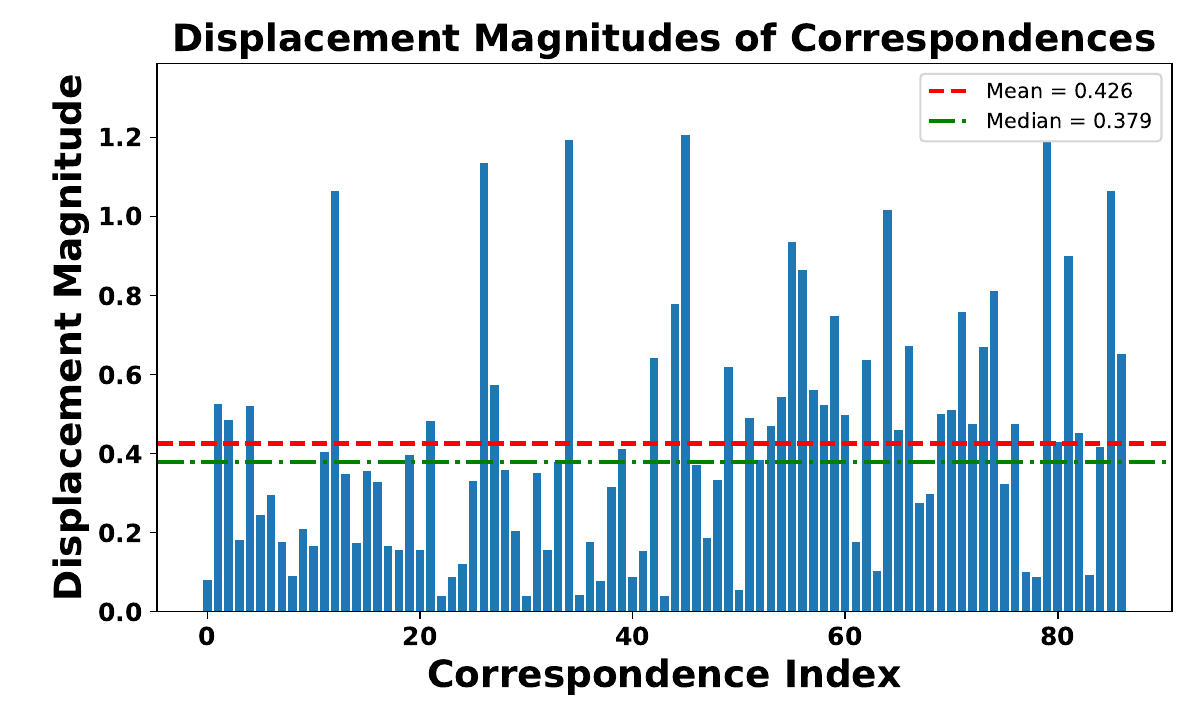} \\
     \includegraphics[width=3.7cm, height=2.3cm]{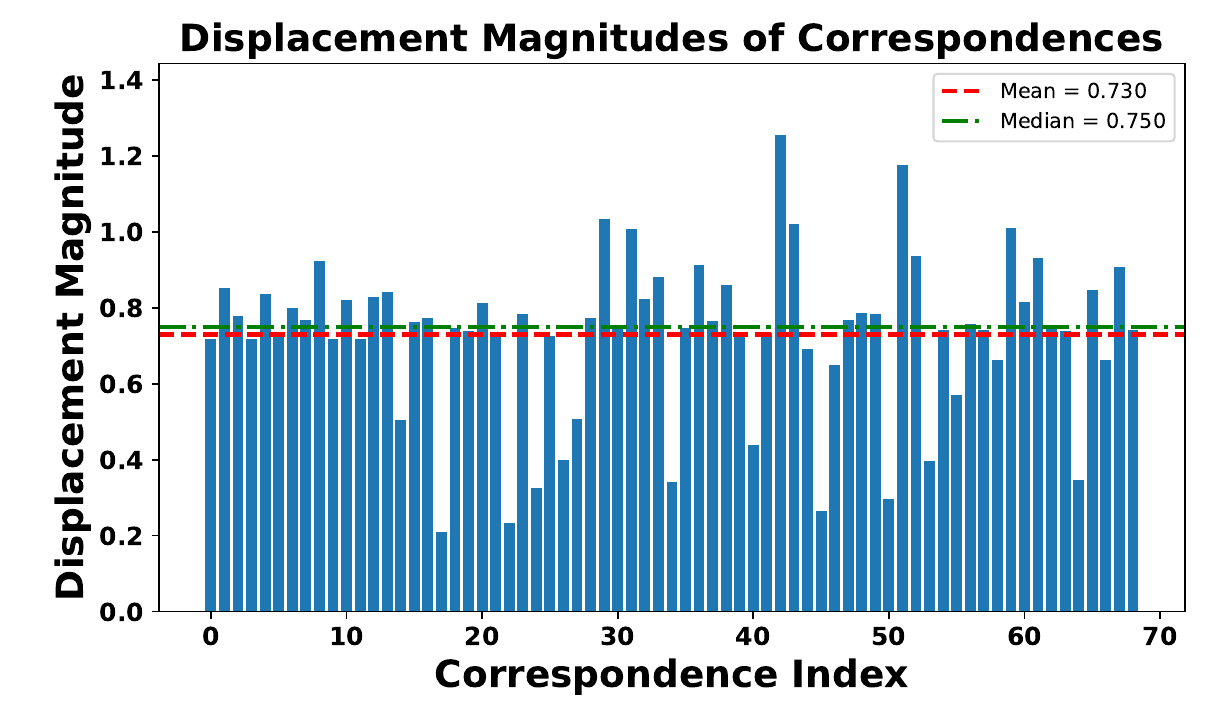} & \includegraphics[width=3.7cm, height=2.3cm]{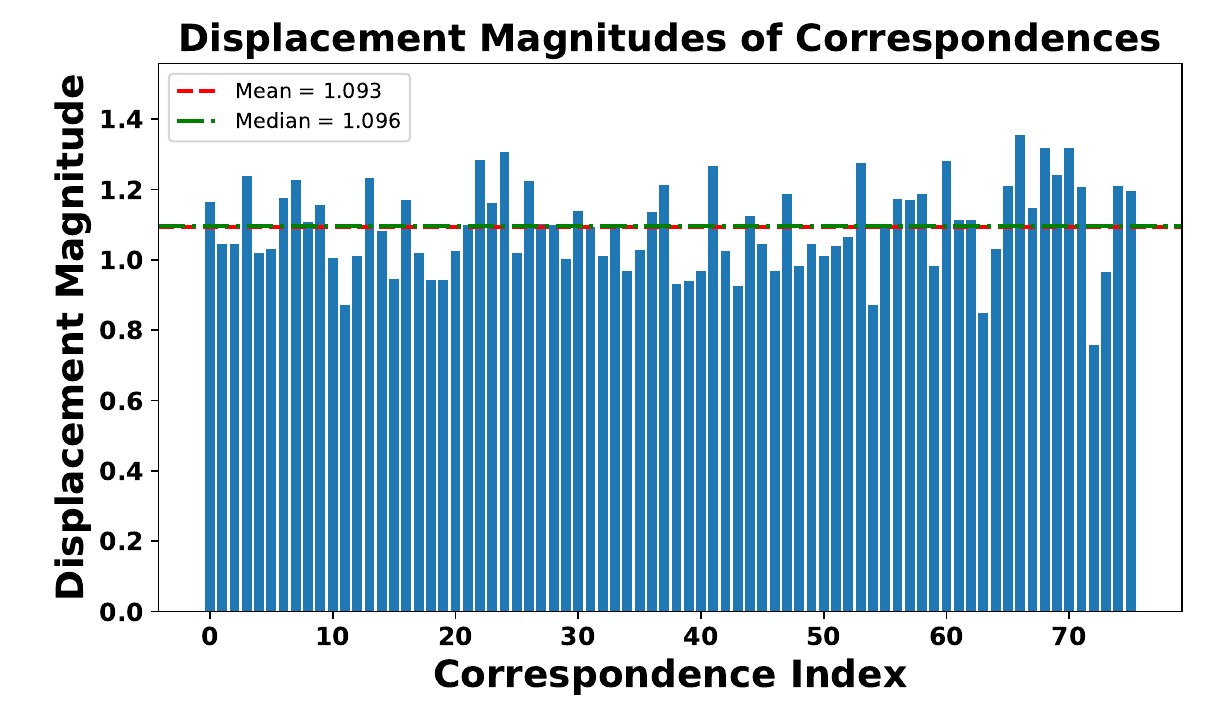} & \includegraphics[width=3.7cm, height=2.3cm]{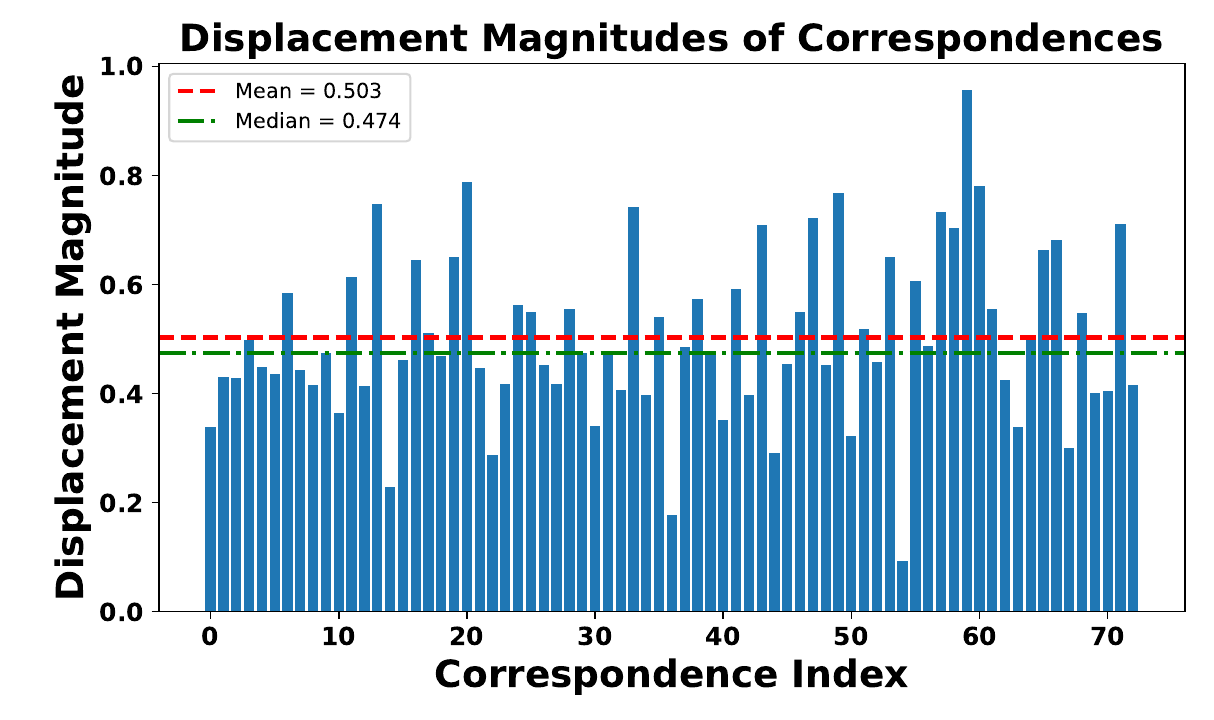} & \includegraphics[width=3.7cm, height=2.3cm]{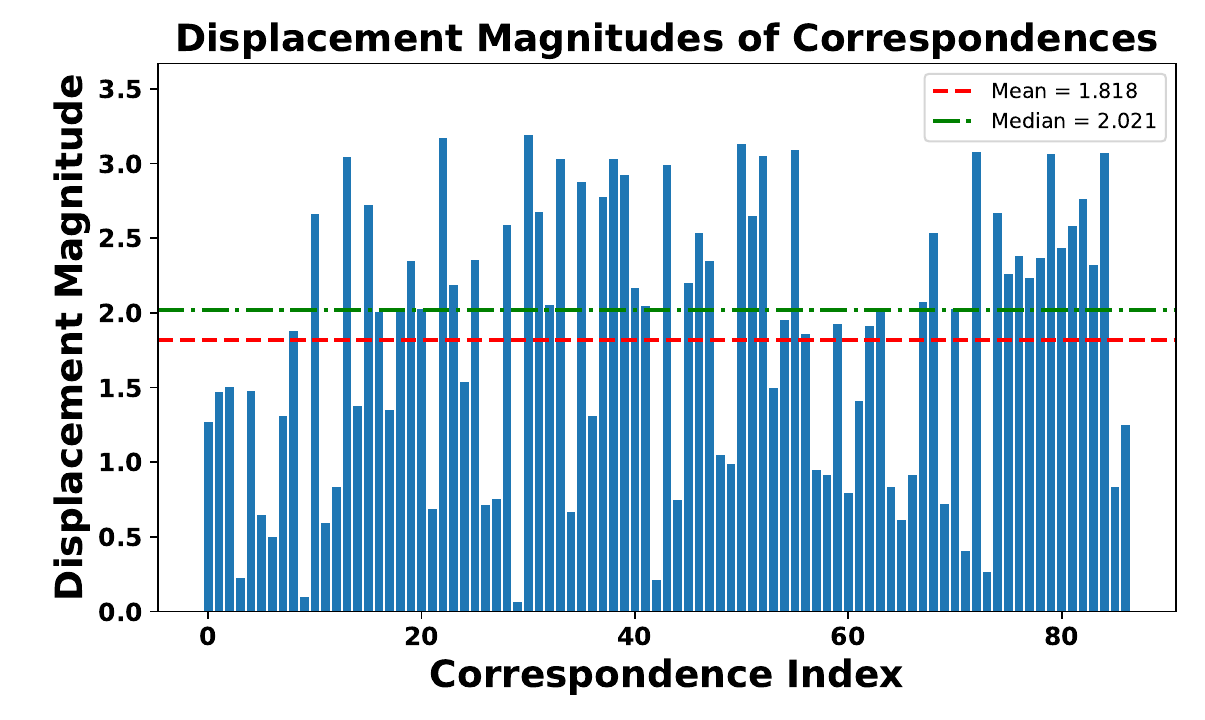} \\
     \includegraphics[width=3.7cm, height=3.7cm]{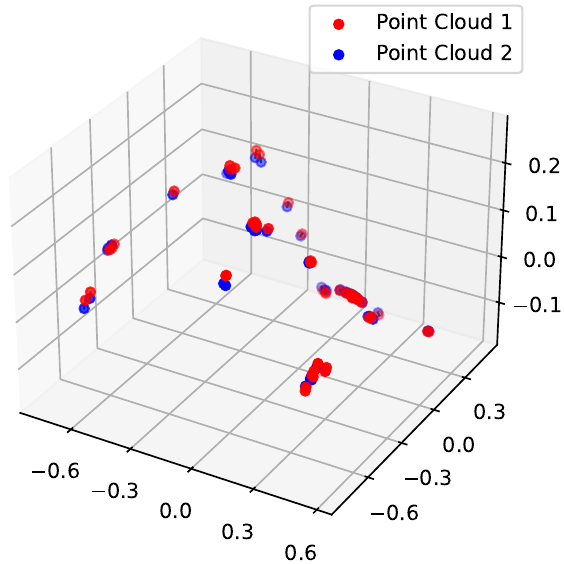} & \includegraphics[width=3.7cm, height=3.7cm]{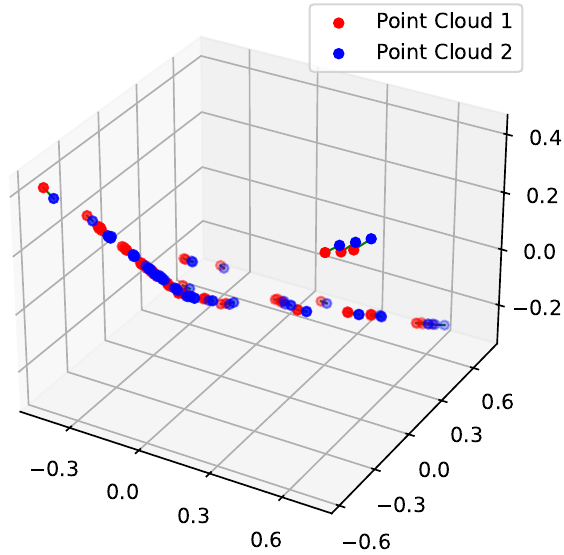} & \includegraphics[width=3.7cm, height=3.7cm]{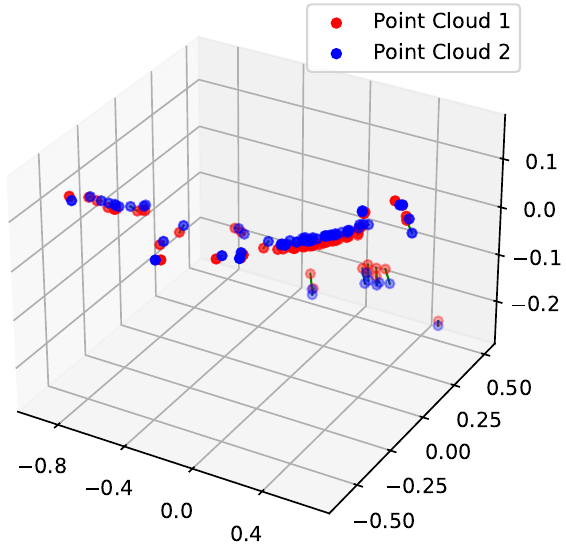} & \includegraphics[width=3.7cm, height=3.7cm]{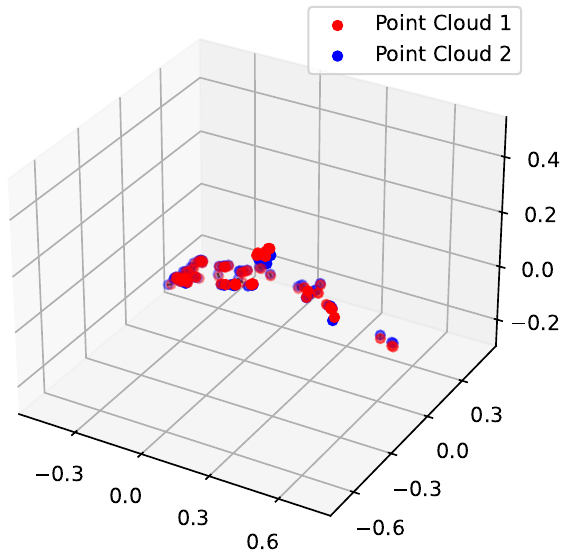} \\
     \includegraphics[width=3.7cm, height=3.7cm]{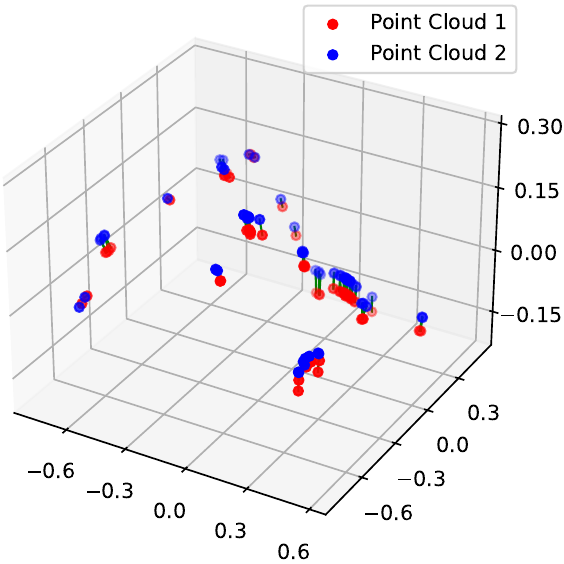} & \includegraphics[width=3.7cm, height=3.7cm]{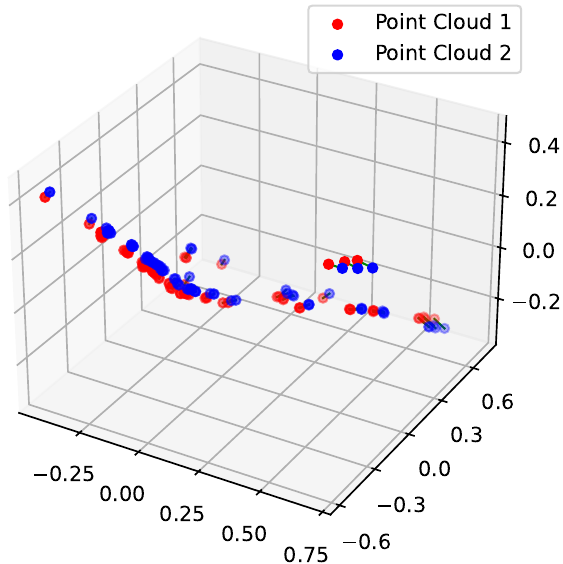} & \includegraphics[width=3.7cm, height=3.7cm]{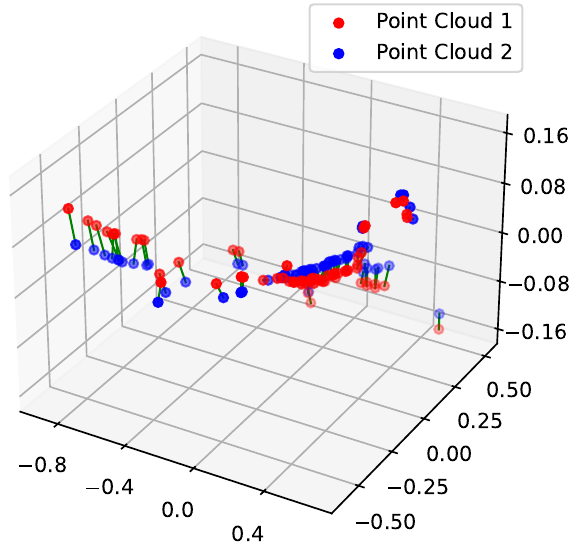} & \includegraphics[width=3.7cm, height=3.7cm]{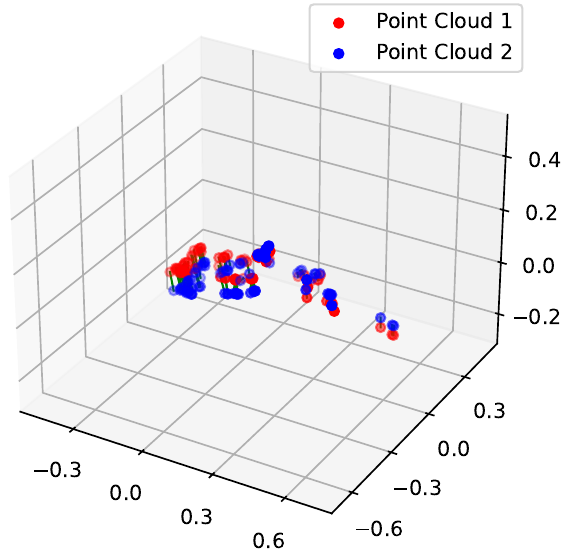} \\
     \end{tabular}
     \caption{Point-to-point correspondence analysis across datasets (a)-(d), matching the scenes in Fig.~\ref{depthPIS0}. First and second rows: displacement magnitudes between matched 3D points derived from MAPIS-Depth and MDEs~\cite{bochkovskii2024depth}, respectively. Third and fourth rows: corresponding 3D scatter plots of matched point pairs. \label{depthPIS00}}
 \end{figure*}

 To quantitatively evaluate the spatial coherence of consecutive RPCs, Fig.~\ref{depthPIS00} presents a point-to-point correspondence analysis between two temporally adjacent point clouds across the datasets shown in Fig.~\ref{depthPIS0}~(a)-(d). For each dataset, ORB features~\cite{rublee2011orb} were extracted from successive RGB frames and projected into \textit{3D} space using the estimated camera poses and intrinsic parameters. These projections were then sampled and paired across frames to compute displacement magnitudes. Each column in Fig.~\ref{depthPIS00} corresponds to a dataset visualised in Fig.~\ref{depthPIS0}. The first and second rows present bar graphs illustrating the displacement magnitudes of the matched point pairs derived from MAPIS-Depth and MDEs~\cite{bochkovskii2024depth}, respectively. The reduced mean and median values in these graphs clearly demonstrate the superior temporal consistency achieved by MAPIS-Depth. The third and fourth rows display the corresponding \textit{3D} scatter plots of the matched point pairs for MAPIS-Depth and MDEs, respectively, providing visual evidence of the alignment quality. As observed, MAPIS-Depth yields lower displacement magnitudes and more tightly clustered spatial alignments, reflecting improved inter-frame consistency and robustness to deformation and motion. In contrast, MDE-based predictions result in larger correspondence errors and scattered point alignments, indicating temporal instability under the dynamic conditions of MIS procedures. These findings quantitatively reinforce the capability of MAPIS-Depth to maintain both temporal coherence across frames, which is an essential requirement for generating reliable \textit{3D} tissue reconstructions in surgical environments.
 
  \par
 The third experiment evaluates the endoscope pose estimation performance of the proposed framework through comparative benchmarking against four state-of-the-art (SOTA) methods: Endo-2DTAM~\cite{huang2025advancing}, OneSLAM~\cite{teufel2024oneslam}, BodySLAM~\cite{henning2022bodyslam}, and Endo-Depth-and-Motion~\cite{recasens2021endo}. The evaluation is performed across representative sequences from the SCARED dataset~\cite{allan2021stereo}, with results summarised in Table~\ref{tab:pose_metrics_combined}. The proposed method consistently outperforms existing approaches in both per-axis and global pose error metrics, including average translation ($T_{\mathrm{avg}}$), average rotation ($R_{\mathrm{avg}}$), and root mean square error. Notably, it also yields lower terminal errors ($T_{\mathrm{fin}}$, $R_{\mathrm{fin}}$) and reduced frame-level drift, as reflected by metrics such as RPE, $T_{\mathrm{max}}$, and $R_{\mathrm{max}}$. These improvements demonstrate the robustness and precision of the proposed approach in handling complex surgical motion and non-rigid deformations, which are critical for reliable endoscope tracking in MMIS procedures.
 \par
 To further substantiate the pose estimation capabilities of the proposed framework, the experiment is extended to a broader comparative evaluation against additional existing methods~\cite{besl1992method, chen1992object, steinbrucker2011real, park2017colored, recasens2021endo}, with results summarised in Table~\ref{Quantitative_comp}. The consistent performance across multiple metrics and baselines affirms its applicability for monocular endoscope tracking in dynamic surgical environments.

   \begin{table*}[htbp]
 \centering
 \caption{Comprehensive pose error metrics reported as mean ± standard deviation. All metrics: lower is better ($\downarrow$). Translation errors are in mm; rotation errors are in degrees (°).}
 \label{tab:pose_metrics_combined}
 \resizebox{\textwidth}{!}{%
 \begin{tabular}{clcccccccc}
 \toprule
 & &
 \multicolumn{2}{c}{\textbf{Per-axis Errors}} &
 \multicolumn{3}{c}{\textbf{Global Errors}} &
 \multicolumn{3}{c}{\textbf{Frame-level Errors}} \\ \hline
 \midrule
 \textbf{Dataset} & \textbf{Framework} & $\mathbf{T_{avg}}$ (mm) \textcolor{blue}{$\downarrow$} & $\mathbf{R_{avg}}$ (°) \textcolor{blue}{$\downarrow$} & $\mathbf{RMSE}$ (mm) \textcolor{blue}{$\downarrow$} & 
 $\mathbf{T_{fin}}$ (mm) \textcolor{blue}{$\downarrow$} & $\mathbf{R_{fin}}$ (°) \textcolor{blue}{$\downarrow$} & $\mathbf{RPE}$ (mm) \textcolor{blue}{$\downarrow$} & 
 $\mathbf{T_{max}}$ (mm) \textcolor{blue}{$\downarrow$} & $\mathbf{R_{max}}$ (°) \textcolor{blue}{$\downarrow$} \\ \midrule
 \midrule
 & \textbf{Proposed}   & $5.556 \pm 3.360$  & $0.061 \pm 0.031$ & $6.493$ & $12.224$ & $0.111$ & $6.220 \pm 3.533$  & $12.236$ & $0.111$ \\
 & \textbf{Endo-2DTAM}\cite{huang2025advancing} & $12.095 \pm 5.782$  & $0.308 \pm 0.152$ & $13.406$ & $26.209$ & $0.573$ & $11.614 \pm 5.878$ & $26.209$ & $0.573$ \\
 \textit{\textbf{S}}\textbf{1} & \textbf{OneSLAM}~\cite{teufel2024oneslam}   & $24.555 \pm 10.315$ & $0.353 \pm 0.174$ & $26.634$ & $40.760$ & $0.470$ & $18.451 \pm 10.005$ & $40.760$ & $0.559$ \\
 & \textbf{BodySLAM}~\cite{henning2022bodyslam}   & $19.147 \pm 8.882$  & $0.536 \pm 0.304$ & $21.106$ & $28.440$ & $0.977$ & $16.352 \pm 8.359$ & $28.440$ & $0.977$ \\
 & \textbf{Endo-Depth-and-Motion}~\cite{recasens2021endo}       & $9.966 \pm 4.534$   & $0.066 \pm 0.032$ & $10.949$ & $16.345$ & $0.106$ & $8.322 \pm 4.497$  & $16.351$ & $0.107$ \\
 \midrule
 & \textbf{Proposed}   & $10.458 \pm 3.028$  & $0.188 \pm 0.083$ & $10.888$ & $11.188$ & $0.093$ & $10.108 \pm 3.191$  & $18.627$ & $0.463$ \\
 & \textbf{Endo-2DTAM}\cite{huang2025advancing} & $ 19.652 \pm 8.199$  & $0.880 \pm 0.570$ & $21.293$ & $18.210$ & $1.432$ & $12.498 \pm 4.115$ & $28.388$ & $3.105$ \\
 \textit{\textbf{S}}\textbf{2} & \textbf{OneSLAM}~\cite{teufel2024oneslam}   & $40.286 \pm 14.969$ & $0.872 \pm 0.379$ & $42.977$ & $32.050$ & $1.057$ & $26.234 \pm 9.530$ & $76.255$ & $1.932$ \\
 & \textbf{BodySLAM}~\cite{henning2022bodyslam}  & $103.702 \pm 48.870$ & $1.650 \pm 0.810$ & $114.640$ & $185.537$ & $3.054$ & $96.247 \pm 50.492$ & $185.537$ & $3.054$ \\
 & \textbf{Endo-Depth-and-Motion}~\cite{recasens2021endo}       & $18.754 \pm 4.629$ & $0.192 \pm 0.050$ & $19.317$ & $17.420$ & $0.194$ & $10.651 \pm 2.518$ & $25.694$ & $0.296$ \\
 \midrule 
 & \textbf{Proposed}   & $23.604 \pm 12.455$  & $0.304 \pm 0.147$ & $26.688$ & $40.247$ & $0.321$ & $31.861 \pm 10.880$  & $47.299$ & $0.530$ \\
 & \textbf{Endo-2DTAM}\cite{huang2025advancing}  & $38.339 \pm 20.230$  & $1.585 \pm 1.025$ & $43.350$ & $74.719$ & $2.040$ & $39.045 \pm 19.130$  & $75.655$ & $3.138$ \\
 \textit{\textbf{S}}\textbf{3} & \textbf{OneSLAM}~\cite{teufel2024oneslam}    & $52.214 \pm 32.595$  & $0.901 \pm 0.451$ & $61.553$ & $49.796$ & $0.978$ & $63.875 \pm 22.562$  & $110.902$ & $1.901$ \\
 & \textbf{BodySLAM}~\cite{henning2022bodyslam}   & $142.554 \pm 92.667$  & $1.332 \pm 0.465$ & $170.026$ & $284.527$ & $2.126$ & $190.421 \pm 95.401$  & $284.527$ & $2.126$ \\
 & \textbf{Endo-Depth-and-Motion}~\cite{recasens2021endo}        & $42.898 \pm 22.935$  & $0.189 \pm 0.087$ & $48.644$ & $70.375$ & $0.366$ & $55.636 \pm 24.131$  & $70.464$ & $0.368$ \\
 \midrule
 & \textbf{Proposed}   & $19.304 \pm 11.368$  & $0.061 \pm 0.031$ & $22.402$ & $42.783$ & $0.482$ & $24.731 \pm 9.424$  & $44.757$ & $0.670$ \\
 & \textbf{Endo-2DTAM}\cite{huang2025advancing}  & $21.187 \pm 9.214$  & $0.931 \pm 0.8311$ & $23.104$ & $29.423$ & $1.229$ & $18.888 \pm 7.646$  & $29.806$ & $3.129$ \\
 \textit{\textbf{S}}\textbf{4} & \textbf{OneSLAM}~\cite{teufel2024oneslam}    & $30.664 \pm 18.135$  & $0.259 \pm 0.139$ & $35.625$ & $53.065$ & $0.479$ & $30.892 \pm 15.196$  & $53.065$ & $0.479$ \\
 & \textbf{BodySLAM}~\cite{henning2022bodyslam}   & $85.127 \pm 43.828$  & $1.866 \pm 0.861$ & $95.747$ & $147.418$ & $2.355$ & $91.727 \pm 33.255$  & $148.072$ & $3.142$ \\
 & \textbf{Endo-Depth-and-Motion}~\cite{recasens2021endo}        & $34.475 \pm 22.319$  & $0.423 \pm 0.259$ & $41.069$ & $72.736$ & $0.781$ & $40.998 \pm 16.859$  & $80.743$ & $0.820$ \\
 \bottomrule 
 \end{tabular}
 }
 \end{table*}

\begin{figure}[ht!]
    \centering
    \begin{minipage}[t]{0.63\linewidth}
        \centering
        \includegraphics[width=\linewidth]{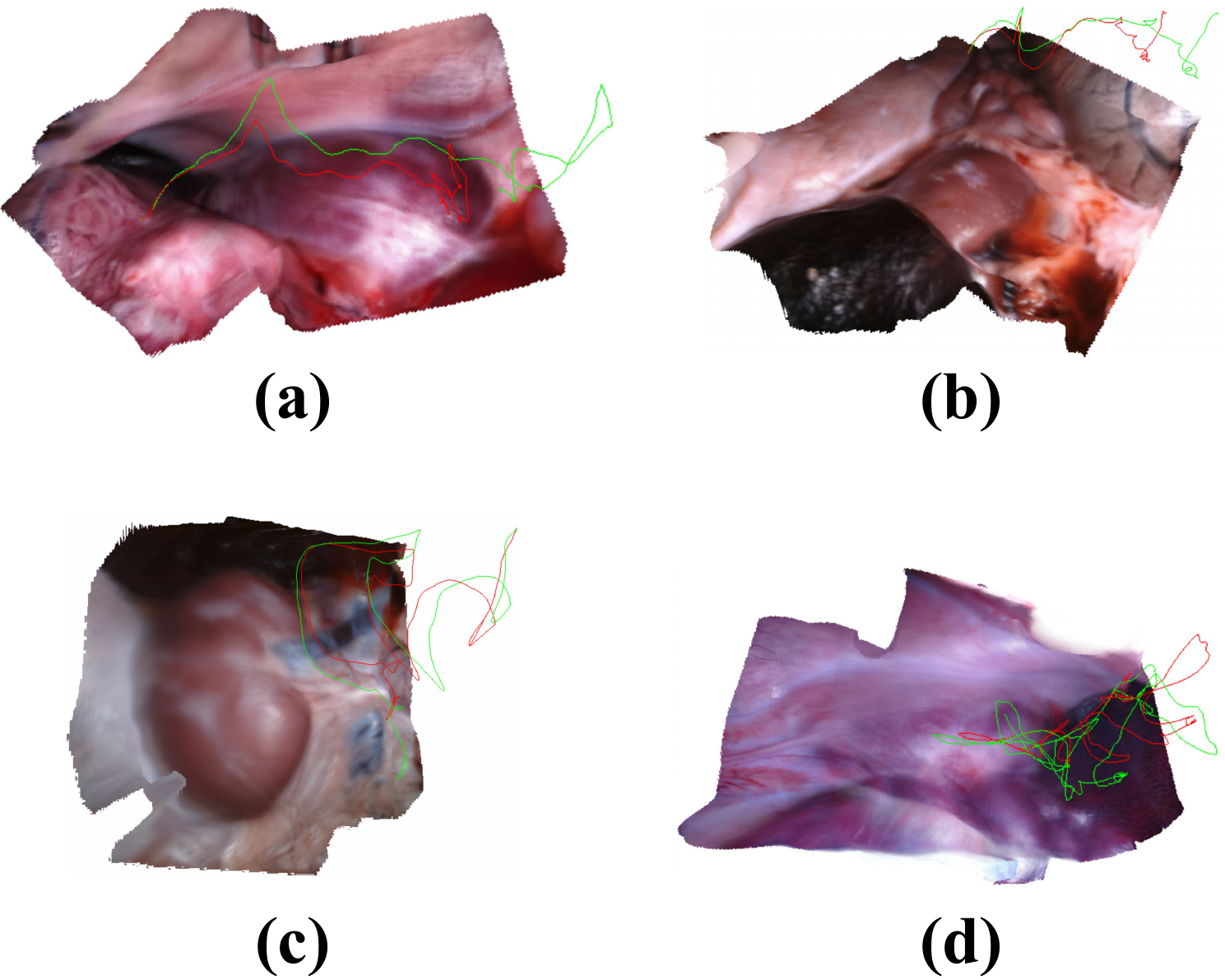}
        \caption{Tissue surface \textit{3D} reconstructions with their corresponding estimated (green) and GT (red) trajectories for SCARED datasets~\cite{allan2021stereo}.}
        \label{reconstructions}
    \end{minipage}%
    \hfill
    \begin{minipage}[t]{0.29\linewidth}
        \centering
        \includegraphics[width=\linewidth]{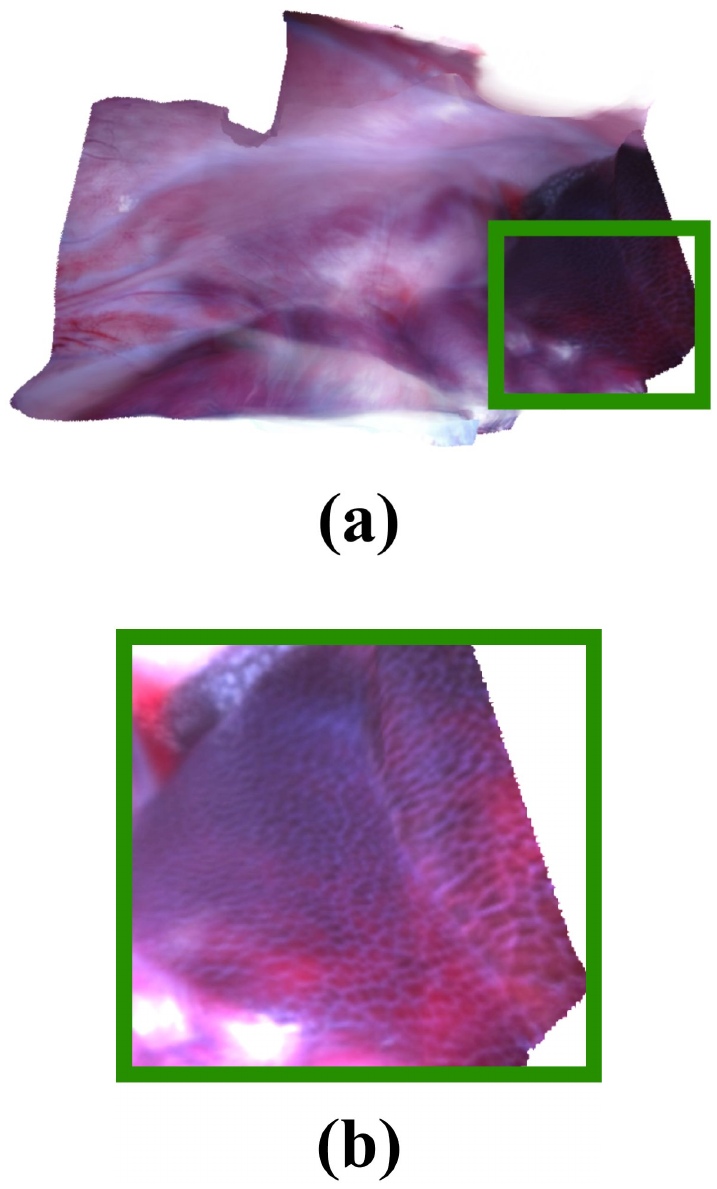}
        \caption{A reconstructed tissue surface with a green rectangle (left) and the rectangle contained region (right) for a SCARED dataset~\cite{allan2021stereo}.}
        \label{reconstructions_zoom}
    \end{minipage}
\end{figure}

  Building upon the previous experiments focused on depth consistency and estimated pose accuracy, the fourth experiment demonstrates the framework’s capability to generate accurate \textit{3D} reconstructions of tissue surfaces in monocular settings. As illustrated in Fig.~\ref{reconstructions}, the method achieves high-fidelity reconstructions while simultaneously estimating the endoscope trajectory across diverse surgical scenes. Despite challenges such as specular reflections, low tissue contrast, steep surface gradients, abrupt motion, and complex anatomical variations, the framework maintains robust spatial and visual consistency. The overlaid estimated (green) and ground truth (red) trajectories further visualise the accuracy of alignment across the reconstructed surfaces.
 \par
 To highlight texture preservation, Fig.~\ref{reconstructions_zoom} presents a localised region of a visually complex surface. Panel~(a) shows a low-contrast, anatomically irregular area, while panel~(b) provides a magnified view of the selected region. The preserved fine-grained textures confirm the framework’s robustness in retaining surface detail and structural coherence, reinforcing its suitability for monocular \textit{3D} reconstruction in endoscopic imaging.

 .

 \begin{table*}
\centering
\caption{Quantitative comparison between the proposed framework and existing methods~\cite{recasens2021endo, besl1992method, chen1992object, steinbrucker2011real, park2017colored} on the SCARED dataset in terms of pose estimation accuracy.}
\label{Quantitative_comp}
\resizebox{\textwidth}{!}{%
\begin{tabular}{lccccccccc}
\toprule
\multicolumn{1}{c}{} & \multicolumn{2}{c}{ \textbf{Per-axis Errors}} & \multicolumn{2}{c}{ \textbf{Global Errors}} & \multicolumn{3}{c}{ \textbf{Frame-level Errors}} & \multicolumn{2}{c}{\textbf{Proportions}}
\\ \cmidrule(lr){2-3} \cmidrule(lr){4-5} \cmidrule(lr){6-8} \cmidrule(lr){9-10}
 \textbf{Framework} & $\mathbf{T_{avg}}$ (mm) \textcolor{blue}{$\downarrow$} & $\mathbf{R_{avg}}$ (°) \textcolor{blue}{$\downarrow$}  & $\mathbf{T_{fin}}$ (mm) \textcolor{blue}{$\downarrow$} & $\mathbf{R_{fin}}$ (°) \textcolor{blue}{$\downarrow$}  & $\mathbf{RPE}$ (mm) \textcolor{blue}{$\downarrow$} & $\mathbf{T_{max}}$ (mm) \textcolor{blue}{$\downarrow$} & $\mathbf{R_{max}}$ (°) \textcolor{blue}{$\downarrow$} & \(\mathbf{T<T_{avg}}\) \textcolor{blue}{\(\uparrow\)}
       & \(\mathbf{R<R_{avg}}\) \textcolor{blue}{\(\uparrow\)}\\
\midrule \midrule
\textbf{Proposed} & $12.551$ & $0.177$ & $15.973$ & $0.342$ & $7.197$ & $18.538$ & $0.347$ & $53.66$ & $50.31$\\ 
\textbf{EDAM}~\cite{recasens2021endo} & $17.320$ & $0.171$ & $22.297$ & $0.300$ & $12.845$ & $26.425$ & $0.281$ & $48.43$ & $46.22$\\ 
\textbf{Park}~\cite{park2017colored} & $17.074$ & $0.206$ & $21.250$ & $0.264$ & $9.713$  & $23.749$ & $0.333$ & $43.73$ & $47.19$\\ 
\textbf{SB}~\cite{steinbrucker2011real} & $16.823$ & $0.197$ & $21.835$ & $0.401$ & $9.962$ & $23.574$ & $0.329$ & $46.76$ & $51.20$\\ 
\textbf{Po2Pl}~\cite{chen1992object} & $19.015$ & $1.958$ & $20.644$ & $2.648$ & $15.423$ & $31.096$ & $3.135$ & $45.12$ & $41.05$\\ 
\textbf{Po2Po}~\cite{besl1992method} & $18.554$ & $1.488$ & $19.314$ & $1.322$ & $15.021$ & $31.140$ & $2.367$ & $46.83$ & $47.50$\\
\midrule
\end{tabular}
}
\end{table*}

 \par
 The fifth experiment, shown in Table~\ref{Ablation}, investigates the influence of key architectural components on scale-awareness through an ablation study using the TLR metric. The proposed framework maintains a TLR of $1.000$ in $S$5, reflecting ideal scale preservation. However, removing Depth Pro and introducing arbitrary scaling factor values, $30$, $100$, and $1000$,  results in considerable TLR deviations from overestimation $1.757$ to severe underestimation $0.052$, underscoring Depth Pro’s essential role in maintaining scale consistency. Similarly, excluding the optical flow module or temporal smoothing components leads to moderate but consistent TLR degradation, confirming their complementary importance. These trends persist across both the SCARED datasets. This study highlights the TLR as a sensitive and interpretable indicator of scale reliability in monocular \textit{3D} reconstruction and validates the necessity of each module for preserving scale consistency in dynamic surgical scenes.

\begin{table*}
\centering
\caption{Ablation study corresponding to SCARED datasets~\cite{allan2021stereo}. The prefix ‘\boldmath{$-$}’ denotes removal of the specified component from the proposed framework, and the scaling factor indicates replacement of Depth Pro initialisation with a fixed multiplier.}
\label{Ablation}
\resizebox{\textwidth}{!}{%
\begin{tabular}{clccccccccccc}
\toprule
\multicolumn{3}{c}{} & \multicolumn{2}{c}{ \textbf{Per-axis Errors}} & \multicolumn{2}{c}{ \textbf{Global Errors}} & \multicolumn{2}{c}{ \textbf{Frame-level Errors}}\\
\cmidrule(lr){4-5} \cmidrule(lr){6-7} \cmidrule(lr){8-9}
Dataset & Framework & \textbf{TLR} & $\mathbf{T_{avg}}$ (mm) \textcolor{blue}{$\downarrow$} & $\mathbf{R_{avg}}$ (°) \textcolor{blue}{$\downarrow$}  & $\mathbf{T_{fin}}$ (mm) \textcolor{blue}{$\downarrow$} & $\mathbf{R_{fin}}$ (°) \textcolor{blue}{$\downarrow$} & $\mathbf{T_{max}}$ (mm) \textcolor{blue}{$\downarrow$} & $\mathbf{R_{max}}$ (°) \textcolor{blue}{$\downarrow$} \\
\midrule
\midrule
 & 
\textbf{Proposed} & $1.000$ & $23.476 \pm 1.409$ & $0.063 \pm 0.036$ & $23.321$ & $0.138$ & $26.132$ & $0.138$ \\
 & \textbf{- Optical Flow} & $1.343$ & $24.574 \pm 0.913$ & $0.062 \pm 0.029$ & $26.281$ & $0.109$ & $26.281$ & $0.109$ \\
 & \textbf{-Depth Pro (scaling factor=30)} & $1.757$ & $24.005 \pm 1.146$ & $0.057 \pm 0.024$ & $26.045$ & $0.111$ & $26.045$ & $0.111$ \\
 \textbf{S}5 & \textbf{-Depth Pro (scaling factor=100)} & $0.520$ & $23.994 \pm 1.157$ & $0.058 \pm 0.025$ & $26.072$ & $0.113$ & $26.072$ & $0.113$ \\
 & \textbf{-Depth Pro (scaling factor=1000)} & $0.052$ & $23.908 \pm 1.180$ & $0.054 \pm 0.022$ & $25.869$ & $0.105$ & $25.979$ & $0.105$ \\
 & \textbf{- EMA} & $1.088$ & $23.969 \pm 1.138$ & $0.063 \pm 0.035$ & $24.780$ & $0.137$ & $26.140$ & $0.137$ \\
 & \textbf{- DyEMA} & $1.247$ & $24.590 \pm 0.992$ & $0.063 \pm 0.032$ & $26.855$ & $0.129$ & $26.855$ & $0.129$ \\
 \midrule
 & \textbf{Proposed} & $0.824$ & $8.111 \pm 3.621$ & $0.457 \pm 0.175$ & $11.519$ & $0.849$ & $16.932$ & $0.219$ \\
 & \textbf{- Optical Flow} & $1.164$ & $12.890 \pm 5.637$ & $0.411 \pm 0.177$ & $22.786$ & $0.706$ & $22.786$ & $0.299$ \\
 & \textbf{-Depth Pro (scaling factor=30)} & $0.990$ & $10.102 \pm 4.172$ & $0.517 \pm 0.209$ & $18.028$ & $0.902$ & $20.092$ & $0.344$ \\
 \textbf{S}6 & \textbf{-Depth Pro (scaling factor=100)} & $0.297$ & $9.787 \pm 4.176$ & $0.506 \pm 0.204$ & $17.380$ & $0.897$ & $19.537$ & $0.344$ \\
 & \textbf{-Depth Pro (scaling factor=1000)} & $0.030$ & $9.857 \pm 4.252$ & $0.507 \pm 0.206$ & $17.578$ & $0.905$ & $19.712$ & $0.344$ \\
 & \textbf{- EMA} & $0.700$ & $7.753 \pm 3.670$ & $0.467 \pm 0.177$ & $8.482$ & $0.847$ & $16.970$ & $0.344$ \\
 & \textbf{- DyEMA} & $0.818$ & $8.116 \pm 3.741$ & $0.467 \pm 0.175$ & $11.082$ & $0.847$ & $17.583$ & $0.344$ \\
\hline
\end{tabular}
}
\end{table*}
 \par
\begin{figure}
    \centering
    \includegraphics[width=1.0\linewidth]{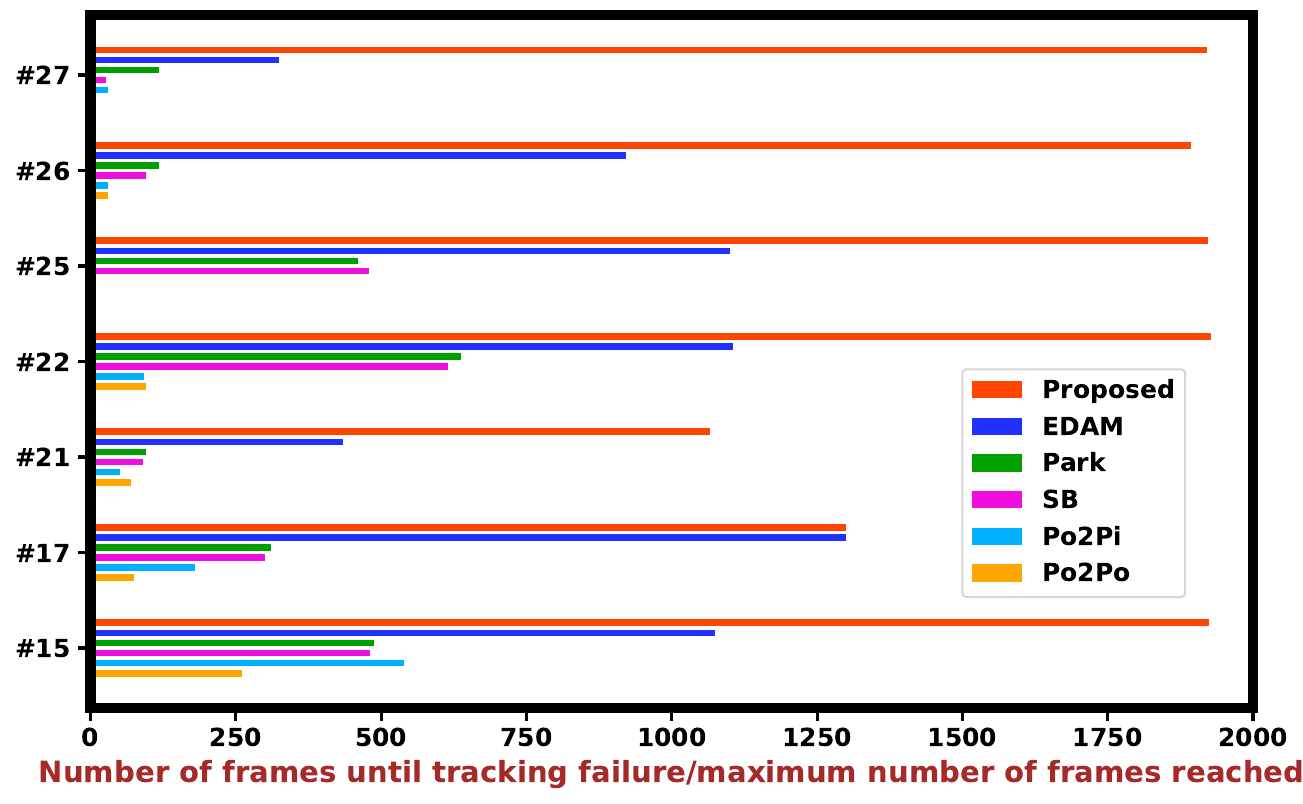}
    \caption{Result comparison between the proposed and existing frameworks~\cite{besl1992method, chen1992object, steinbrucker2011real, park2017colored, recasens2021endo} in terms of tracking convergence failures on various HEVD datasets~\cite{recasens2021endo} as shown along the vertical axis.}
    \label{convergence}
\end{figure}

 The final experiment, shown in Fig.~\ref{convergence}, compares the convergence stability of the proposed framework with existing methods~\cite{besl1992method, chen1992object, steinbrucker2011real, park2017colored, recasens2021endo} using the HEVD dataset~\cite{recasens2021endo}. Convergence is assessed based on the number of frames processed before tracking failure occurs. The proposed framework consistently maintains trajectory estimation across the full length of each sequence, reaching the maximum number of frames without failure. In contrast, competing approaches frequently terminate prematurely. These results highlight the framework’s robustness in maintaining long-term stability and resilience under challenging monocular conditions.

 \section{Conclusion}

 This work presents a unified framework for monocular endoscope pose estimation and tissue surface reconstruction that addresses depth ambiguity, physiological tissue deformation, inconsistent endoscope motion, limited texture fidelity, and a restricted field of view. The integration of MAPIS-Depth, leveraging Depth Pro for initialisation and Depth Anything for per-frame prediction provides scale-consistent depth estimates. These are temporally refined through RAFT-based correspondence and LPIPS-guided perceptual similarity, mitigating artefacts due to motion and deformation. The WEMA-RTDL module ensures robust pose estimation, and TSDF-based volumetric fusion with marching cubes yields coherent \textit{3D} surface meshes. Extensive experimental validation across multiple endoscopic datasets confirms the framework’s superiority in tissue reconstruction, demonstrating improved depth consistency, precise endoscope pose estimation, and enhanced scale awareness compared to existing approaches. Future work may explore uncertainty-guided refinement and multimodal integration, extending its applicability to real-time robotic surgery and intraoperative decision support.

 \bibliographystyle{unsrt} 
 \bibliography{main.bib}

\end{document}